\documentclass[10pt,twocolumn,letterpaper]{article}

\usepackage[pagenumbers]{cvpr} % To force page numbers, e.g. for an arXiv version

\definecolor{cvprblue}{rgb}{0.21,0.49,0.74}
\usepackage[pagebackref,breaklinks,colorlinks,allcolors=cvprblue]{hyperref}

\usepackage{multirow, booktabs}
\usepackage{stfloats}
\usepackage{graphicx}
\usepackage{caption}
\usepackage{algorithm}
\usepackage{algorithmic}
\usepackage{soul}

\def\paperID{***} %
\def\confName{***}
\def\confYear{***}

\title{FDS: Frequency-Aware Denoising Score\\ for Text-Guided Latent Diffusion Image Editing}

\author{
    Yufan Ren$^1$ \quad
    Zicong Jiang$^{1,2,3}$\quad
    Tong Zhang$^{1}$\href{mailto:tozhang.ucas@gmail.com}{\textsuperscript{†}} \quad
    Søren Forchhammer$^2$ \quad 
    Sabine Süsstrunk$^1$ \\
    $^1$School of Computer and Communication Sciences, EPFL\\ $^2$Department of Electrical and Photonics Engineering, DTU 
    \\ $^3$ Department of Electrical Engineering, Chalmers University of Technology 
}

\begin{document}

\twocolumn[{
\begin{center}
   \maketitle
   \captionsetup{type=figure}\includegraphics[width=1.0\textwidth,page=1]{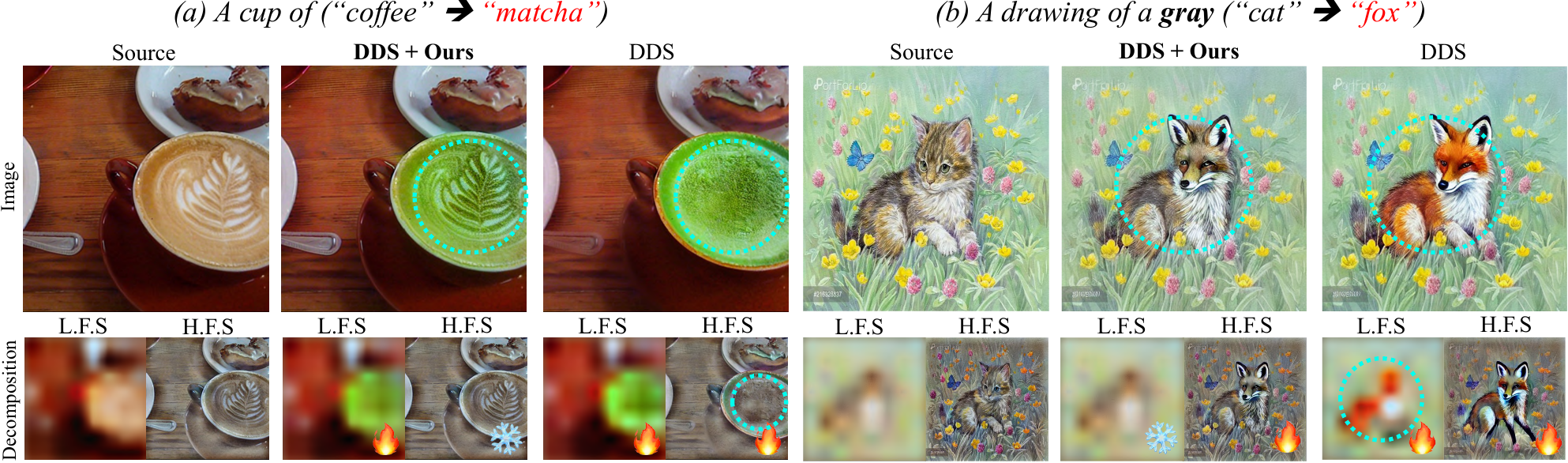} 
    \captionof{figure}{Text-guided image editing using Text-to-Image (T2I) models, such as DDS~\cite{hertz2023delta}, often fails to produce satisfactory results due to \textbf{indiscriminate optimization across all frequency subbands}. For example, in the top row, DDS removes the detailed pattern of latte art (a) and drastically alters the cat's color despite ``gray" being specified in the prompt (b). 
    These issues become more apparent through frequency decomposition during optimization (L.F.S and H.F.S refer to low frequency subband and high frequency subband, respectively) in the second row, where unnecessary modifications occur. Our method selectively optimizes frequency bands, preserving high-frequency details in the latte art (a) and maintaining color consistency in the gray cat (b). The ``freeze" symbol indicates frozen frequency components, while the ``flame" means optimized. Best viewed on a screen when zoomed in.
    }
    \label{fig:teaser_first_page}
\end{center}
}]
\renewcommand{\thefootnote}{} 
\footnote{%
\parbox{\linewidth}{%
\noindent\href{mailto:tozhang.ucas@gmail.com}{\textsuperscript{†}:}~Corresponding author.
}}

\begin{abstract} 
Text-guided image editing using Text-to-Image (T2I) models often fails to yield satisfactory results, frequently introducing unintended modifications, such as the loss of local detail and color changes. In this paper, we analyze these failure cases and attribute them to the indiscriminate optimization across all frequency bands, even though only specific frequencies may require adjustment. To address this, we introduce a simple yet effective approach that enables the selective optimization of specific frequency bands within localized spatial regions for precise edits. Our method leverages wavelets to decompose images into different spatial resolutions across multiple frequency bands, enabling precise modifications at various levels of detail. To extend the applicability of our approach, we provide a comparative analysis of different frequency-domain techniques. Additionally, we extend our method to 3D texture editing by performing frequency decomposition on the triplane representation, enabling frequency-aware adjustments for 3D textures. Quantitative evaluations and user studies demonstrate the effectiveness of our method in producing high-quality and precise edits. Further details are available on our project website: \href{https://ivrl.github.io/fds-webpage/}{https://ivrl.github.io/fds-webpage/}
\end{abstract}

\section{Introduction}

Recent advances in Text-to-Image (T2I) models, such as DALL-E 2~\cite{ramesh2022hierarchical} and Latent Diffusion Models (LDMs)~\cite{rombach2021highresolution}, have significantly advanced the field of text-based image generation. 
Despite these advances, achieving precise text-guided image editing with T2I models remains challenging. Existing methods, including DreamFusion~\cite{poole2022dreamfusion} with its Score Distillation Sampling (SDS) technique and subsequent enhancements such as Delta Denoising Score (DDS)~\cite{hertz2023delta} and Contrastive Distillation Score (CDS)~\cite{nam2024contrastive}, have made notable progress. However, they still struggle with fine-grained control over edits. In particular, they often fail to preserve high-frequency details or alter the color, leading to unsatisfactory results in tasks that require precise modifications. 
For example, as shown in Fig.~\ref{fig:teaser_first_page}, when DDS is used to transform ``a cup of \textbf{coffee}" into ``a cup of \textbf{matcha}", intricate latte art details are often unintentionally removed; converting a drawing of ``a gray cat" into ``a gray fox" changes the cat's color to reddish-brown, likely due to data bias, despite specifying the color ``\textbf{gray}" in the prompt.

We identify that these limitations arise from two primary challenges. First, language-vision models struggle to disentangle complex attributes~\cite{berzak2016you,mehrabi2023resolving,prasad2023rephrase}, which hinders the faithful preservation of source image attributes that should remain unchanged (e.g., the fox's color in Fig.~\ref{fig:teaser_first_page}). However, fundamentally improving this aspect is resource-intensive and may require retraining a more robust text-image model with better annotated paired data. Second, and more directly, T2I editing models are hindered by indiscriminate optimization across all frequency subbands during editing, often causing unintended alterations (both cases in Fig.~\ref{fig:teaser_first_page}).
Therefore, providing additional levels of controllability in the frequency domain for text-guided editing is needed.

In this work, we propose a simple yet effective approach that enhances text-guided image editing by revisiting classical frequency domain techniques. 
We introduce a method that decomposes images into frequency subbands~\cite{daubechies1993ten}, enabling selective optimization of each. We showcase the benefit of frequency-awareness for editing in two cases: local detail preservation and color fidelity preservation. This selective optimization provides more precise control over the editing process and leads to more satisfying results. Additionally, we extend our method to 3D texture editing and propose a frequency-aware texture editing method that uses a frequency-decomposed triplane representation~\cite{Chan2021}. This extension enhances frequency awareness, improving both color fidelity and the preservation of high-frequency details in textures. Our contributions are as follows:

\begin{itemize} 
\item We identify issues with local detail preservation and color fidelity in score distillation-based image editing methods, and attribute these issues to indiscriminate optimization across all frequency subbands. 
\item We propose a novel text-guided image editing method that leverages frequency domain techniques to provide precise control within the frequency domain. 
\item We demonstrate that selectively optimizing specific frequency subbands leads to better preservation of image details and color fidelity compared to baseline methods.
\item We introduce a frequency-decomposed triplane representation tailored for 3D texture editing. 
\item We provide an analysis of different frequency domain techniques and offer a guideline on their application to the text-guided image editing task. 
\end{itemize}

\section{Related Work}
\label{sec:related}

\noindent\textbf{Text-to-Image Models.} Text-to-Image (T2I) models~\cite{karras2022elucidating}, such as DALL-E 2~\cite{ramesh2022hierarchical}, Stable Diffusion~\cite{rombach2021highresolution}, and Imagen~\cite{saharia2022photorealistic}, have seen remarkable success in generating high-quality images from text prompts. Due to their powerful diffusion-based generative priors, T2I models have been widely applied in various domains including depth estimation~\cite{ke2024repurposing}, image-to-image translation~\cite{meng2021sdedit,couairon2022diffedit}, image super-resolution~\cite{wang2024exploiting,moser2024diffusion}, and image inpainting~\cite{corneanu2024latentpaint,pan2024coherent,xie2023smartbrush}. 

\noindent\textbf{Image Editing with Text-to-Image Models}. Recent research has explored various applications of T2I models in image editing, including interacting with the reverse process of an inverted latent~\cite{miyake2023negative,mokady2023null,cho2024noise,ju2023direct,cao2023masactrl,kawar2023imagic,koo2024flexiedit,kim2025dreamsampler,hertz2022prompt} or fine-tuning the diffusion model to learn the reverse process to restore the edited image, such as Instruct-Pix2Pix~\cite{brooks2023instructpix2pix}.
While many of these methods exploit the reverse process of diffusion models, a different
method known as Score Distillation Sampling (SDS)~\cite{poole2022dreamfusion,koo2024posterior}, originally proposed for 3D object generation, has shown promising performance. However, despite SDS being effective in generating 3D assets, its gradient updates can introduce noise and result in imprecise edits, particularly when applied to fine image editing tasks. Delta Denoising Score (DDS)~\cite{hertz2023delta} is a method that improves upon SDS by canceling out noisy gradients and focusing on the desired modifications in a more localized manner, leading to better preservation of image details during editing. CDS~\cite{nam2024contrastive} leverages a constrastive loss term to enforce that the semantics at each spatial position remain the same during optimization. 
 
Our approach is compatible with score distillation-based methods, integrating wavelet decomposition as an image representation to enable selective modification of certain frequency components during editing.

\begin{figure*}[t] \centering \includegraphics[width=1.0\textwidth, page=2]{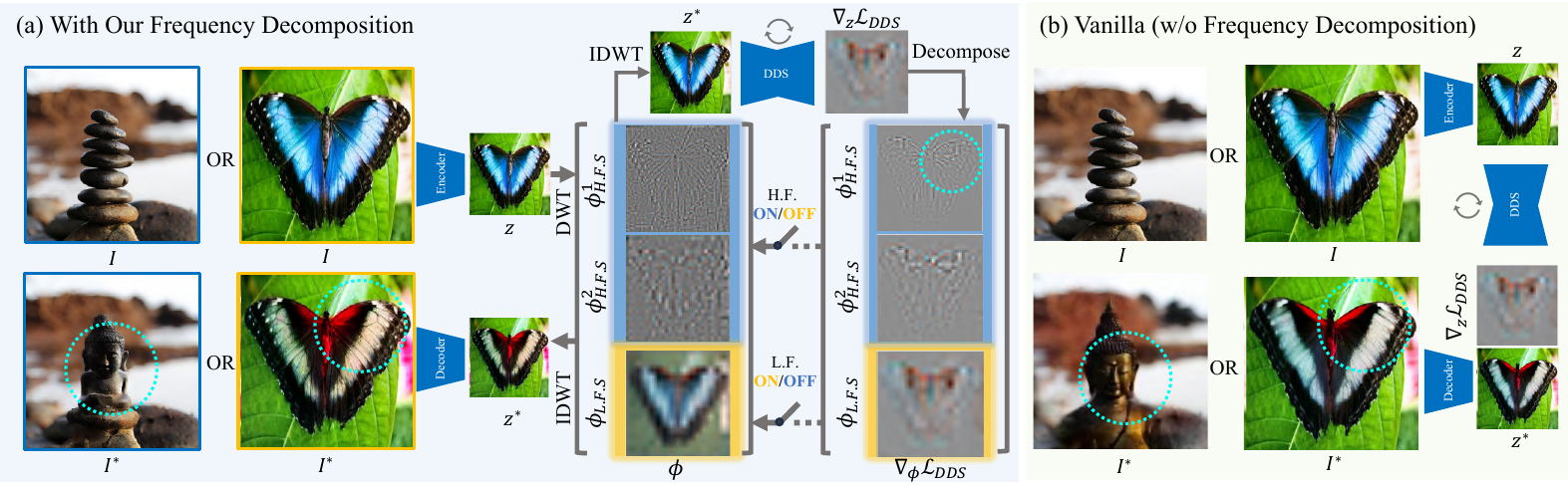} \caption{\textbf{Method overview.} Ours (a) differs from vanilla score distillation editing (b), which backpropagates gradients to the latent space ($z$)\protect\footnotemark to perform editing. Our method leverages wavelet frequency decomposition to decompose latent $z$ into wavelet subbands $\phi$ including high frequency ($\phi_{\text{H.F.S}} = \{\phi_{\text{H.F.S}}^1, \phi_{\text{H.F.S}}^2, \cdots, \phi_{\text{H.F.S}}^J\}$) and low frequency ($\phi_{\text{L.F.S}}$). We process the reconstructed latent $z^*$ with the diffusion model to obtain a gradient for optimization, which is applied to either high-frequency components or low frequency components selectively depending on application. Consequently, our method produces edits that benefit from detail preservation (butterfly case, \textcolor[rgb]{1,0.8,0}{\textbf{yellow}} for text and image borders) and color fidelity (stone, \textcolor{blue}{\textbf{blue}} for text and image borders). Best viewed on a screen when zoomed in.} \label{fig:method} \end{figure*}

\noindent \textbf{Frequency Domain Techniques in Image Editing}. Despite frequency domain techniques having been deployed in various ways of image manipulation~\cite{yang2010medical,krishnamoorthi2009image,drori2003fragment,cai2021frequency,yu2022frequency,yang2022elegant,everaert2024exploiting}, their application to text-guided image editing is still limited. FlexiEdit~\cite{koo2024flexiedit} adds randomness to high frequency components during diffusion inversion allowing for non-rigid editing. FreeDiff~\cite{wu2025freediff} emphasizes high frequency component during the diffusion sampling process.
\cite{gao2024frequency} maps frequency information to images using a ControlNet~\cite{zhang2023adding} fashion.  

Differently, our method leverages Wavelet Transforms~\cite{daubechies1993ten} in fine-grained image editing with score distillation sampling, enabling the preservation of detail and color fidelity. Besides, similar to other score distillation methods, our method does not require training or finetuning diffusion models.

\section{Frequency-Aware Score Distillation}
\label{sec:method}

Our work enhances SDS~\cite{poole2022dreamfusion} and its improved versions, e.g., DDS~\cite{hertz2023delta} and CDS~\cite{nam2024contrastive}, by incorporating Wavelet Transform~\cite{daubechies1993ten} to enable selective frequency optimization. Unlike the purely spatial approach used in standard representations, our method employs a spatial-frequency representation and seamlessly integrates within the score distillation frameworks. Without loss of generality, we demonstrate our approach in image editing based on DDS and texture editing based on SDS as they achieve the best results.

\footnotetext{For latent visualization, we use a linear transform from four-channel latent to three-channel RGB.}

\subsection{Frequency-Decomposed Image Editing}

\subsubsection{Initial Setup and DDS}

Given a source image \( I \) of shape \( 3 \times 512 \times 512 \), we encode it via a Variational Autoencoder (VAE)~\cite{kingma2013auto} to obtain its latent representation \( z \) with dimensions \( 4 \times 64 \times 64 \). In the original DDS approach, optimization is performed directly on \( z \) by calculating the loss gradient \( \nabla_z \mathcal{L}_{\text{DDS}} \) (conditioned on a source prompt and a target prompt) and iteratively updating \( z \) over \( N \) steps before decoding it at the end. 

\subsubsection{Wavelet Decomposition and Subband Extraction}

Unlike the vanilla DDS, we enable selective optimization by decomposing the latent \( z \) into multiple frequency subbands \( \phi \) through DWT~\cite{daubechies1993ten}. We set the decomposition level \( J \) and select an appropriate wavelet type, such as Daubechies wavelets~\cite{daubechies1993ten}, which will be analyzed in Sec.~\ref{sec:discussion}. The decomposition yields subbands \( \phi \), consisting of a low-frequency component \( \phi_{\text{L.F.S}} \) and multiple high-frequency components $\phi_{\text{H.F.S}} = \{ \phi^1_{\text{H.F.S}}, \phi^2_{\text{H.F.S}}, \dots, \phi^J_{\text{H.F.S}} \}$, where each level of high-frequency subbands contains three directional details, such as \( \phi^1_{\text{H.F.S}_{HL}} \), \( \phi^1_{\text{H.F.S}_{LH}} \), and \( \phi^1_{\text{H.F.S}_{HH}} \). The low-frequency subband \( \phi_{\text{L.F.S}} \) has dimensions \( 4 \times \frac{64}{2^J} \times \frac{64}{2^J} \). 

\subsubsection{Selective Optimization}

Using the Inverse Discrete Wavelet Transform (IDWT), we reconstruct $\hat{z}$ from the subbands $\phi$ which is then input into the diffusion model to compute the gradient $\nabla_{\phi} \mathcal{L}_{\text{DDS}}$, which aims to modify the target attributes.

The gradient $\nabla_{\phi} \mathcal{L}_{\text{DDS}}$ propagates back to each subband, including both $\phi_{\text{L.F.S}}$ and $\phi_{\text{H.F.S}}$.
As these subbands are categorized into low-frequency components $\phi_{\text{L.F.S}}$ and high-frequency components $\phi_{\text{H.F.S}}$, we use a stop-gradient operation during optimization (denoted as $\text{StopGrad}(\cdot)$) for selective gradient control.
If the objective is to preserve low-frequency content (e.g., color), we apply the $\text{StopGrad}$ to $\phi_{\text{L.F.S}}$ to prevent updating to it and focusing optimization on $\phi_{\text{H.F.S}}$.
Conversely, if preserving high-frequency details is desired, we apply $\text{StopGrad}$ to $\phi_{\text{H.F.S}}$ to restrict updates to the low-frequency components only.
After $N$ optimization iterations, we decode the optimized latent $\hat{z}$ to produce the final output image.
This output image preserves the original attributes intended for retention, with targeted adjustments as specified by the prompts.

\begin{figure}[t]
\setlength{\belowcaptionskip}{-0.375cm}
    \centering    \includegraphics[width=0.49\textwidth,page=3]{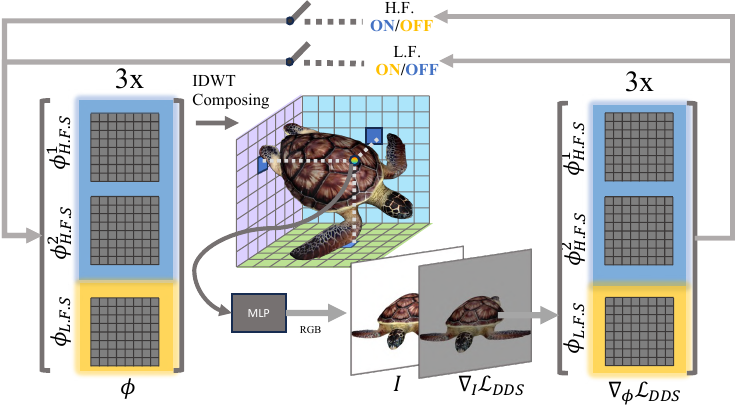}
    \caption{\textbf{3D texture editing pipeline}. We represent a 3D texture field as a frequency-decomposed triplane $\phi$, i.e., three sets of wavelet subbands representing $XY-YZ-XZ$ in three directions. To render an image at camera view $p$, we construct a triplane from $\phi$, which is queried for colors. The rendered image is processed by the latent diffusion model to produce a gradient, which is backpropagated to update selected frequency components.} 
\label{fig:triplane}
\end{figure}

\subsection{Frequency-Aware 3D Texture Editing}

We extend our method to 3D texture editing on meshes, aiming to modify textures while preserving detail and color fidelity (Fig.~\ref{fig:triplane}). 

\subsubsection{Frequency-Aware Triplane Representation}

Given a mesh $\mathbf{M}$ with triangular faces, we denote its original texture before editing as $\mathbf{T}$.
To represent the texture field to be optimized, we use a triplane~\cite{chan2022efficient} with three planes $P_{xy}$, $P_{xz}$, and $P_{zy}$, each sized $C \times H \times W$.
Rendering an image from the mesh  $\mathbf{M}$ requires retrieving the color of each pixel. 
To do so, for each pixel, we cast a ray from the camera and calculate the ray-mesh intersection $\mathbf{x} = [x, y, z]$ (where $x, y, z \in [-1, +1]$ in a normalized space). 
To retrieve the color at this point $\mathbf{x}$,
we project it onto these planes using bilinear interpolation, which yields three $C$-channel vectors.
These three vectors are combined to form a $3 \times C$ feature vector, which is then decoded into RGB values using an MLP.

To incorporate frequency awareness, similar to 2D image editing, we represent each plane in the triplane using a set of wavelet subbands, Fig.~\ref{fig:triplane}.
Before rendering, we apply the IDWT to reconstruct the triplane from subbands, i.e., $\hat P=f_\text{IDWT}(\phi_{\text{H.F.S}}, \phi_{\text{L.F.S}})$ for each dimension and use the reconstructed triplane $\hat P$ in the rendering process described earlier. 
Before editing, we initialize the triplane and MLP by fitting the rendered image to the original texture map $\mathbf{T}$ using the L1 loss $\mathcal{L}_{1}$.

\subsubsection{Selective Optimization}

After rendering the image $I$ from frequency-aware triplane, we use a VAE to encode it into latent $z$. We then apply a diffusion model to compute the score distillation loss $\mathcal{L}_{\text{SDS}}$ and we obtain the gradient $\nabla_{\phi} \mathcal{L}_{\text{SDS}}$ by backpropagation. Similar to 2D image editing, we apply $\text{StopGrad}(\cdot)$ operator to either the low- or high-frequency components.

\begin{figure*}[t] \centering \includegraphics[width=0.95\textwidth, page=4]{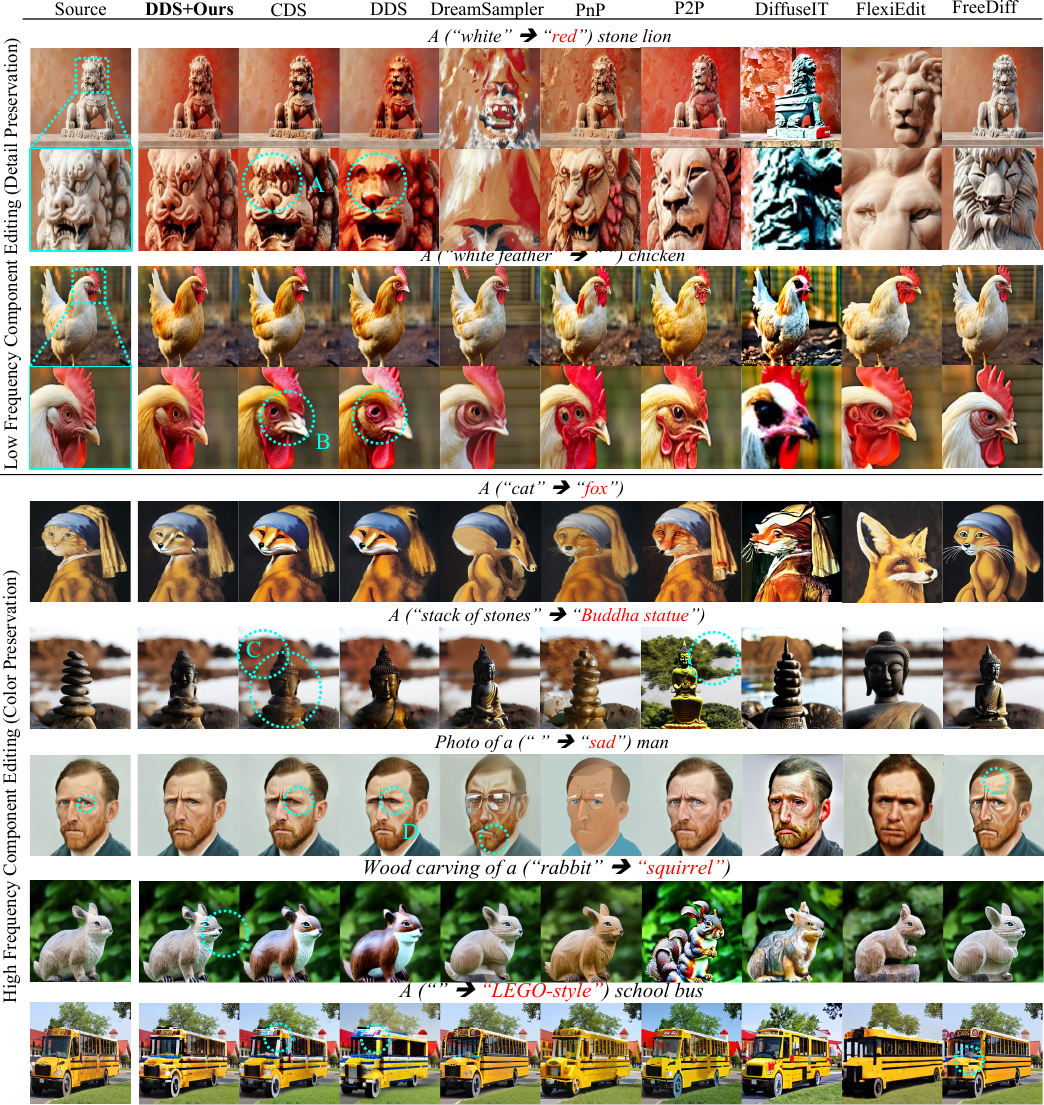} \caption{\textbf{Qualitative results.} We conducted a qualitative comparison with the most competitive baselines. For low-frequency editing, our method follows instructions closely while preserving high-frequency details better. In the first row, stone lion, our method preserves details of the lion's eyes and mouth (\textbf{A}). On the contrary, CDS, DDS and other methods lose these structures, introducing significant changes. For the second row, the chicken, we preserve the beak and eye areas (\textbf{B}). In contrast, other methods distort the structure noticeably or fail to generate meaningful images (e.g., DiffuseIT) and follow the target description (DreamSampler, FlexiEdit). CDS, the best among baselines, alters the beak. For high-frequency editing, our method maintains better color fidelity than the baselines. In the first row, our approach preserves color consistency in the transformation from cat to fox. In the stone-to-Buddha case, our method preserves both the background and statue colors (\textbf{C}) better than CDS and similar methods. In the third row, our method preserves image color information better, especially the pupil and face skin color, while still modifying details (\textbf{D}). Other methods introduce structure distortion, which can be attributed to the lack of global information guidance. Best viewed on a screen when zoomed in.} \label{fig:qualitative:2d} \end{figure*}

\subsection{Implementation Details}

Our implementation leverages PyTorch~\cite{NEURIPS2019_9015}, with wavelet transformations handled by the differentiable wavelet library \texttt{wavelet\_pytorch}~\cite{cotter2020uses}. Similar to \cite{hertz2023delta}, we use Stable Diffusion~\cite{rombach2021highresolution} as the image diffusion model. Moreover, we conduct experiments on an NVIDIA RTX 3090 GPU. We set the number of iterations to $N=500$ for score distillation sampling. 

\section{Experiments}

In this section, we evaluate the effectiveness of our method in both 2D image editing and 3D texture editing. We begin by introducing our baseline methods and experimental protocols. Next, we report qualitative and quantitative comparisons. Finally, we present the results from a user study.

\begin{figure*}[t] \centering \includegraphics[width=1.0\textwidth, page=5]{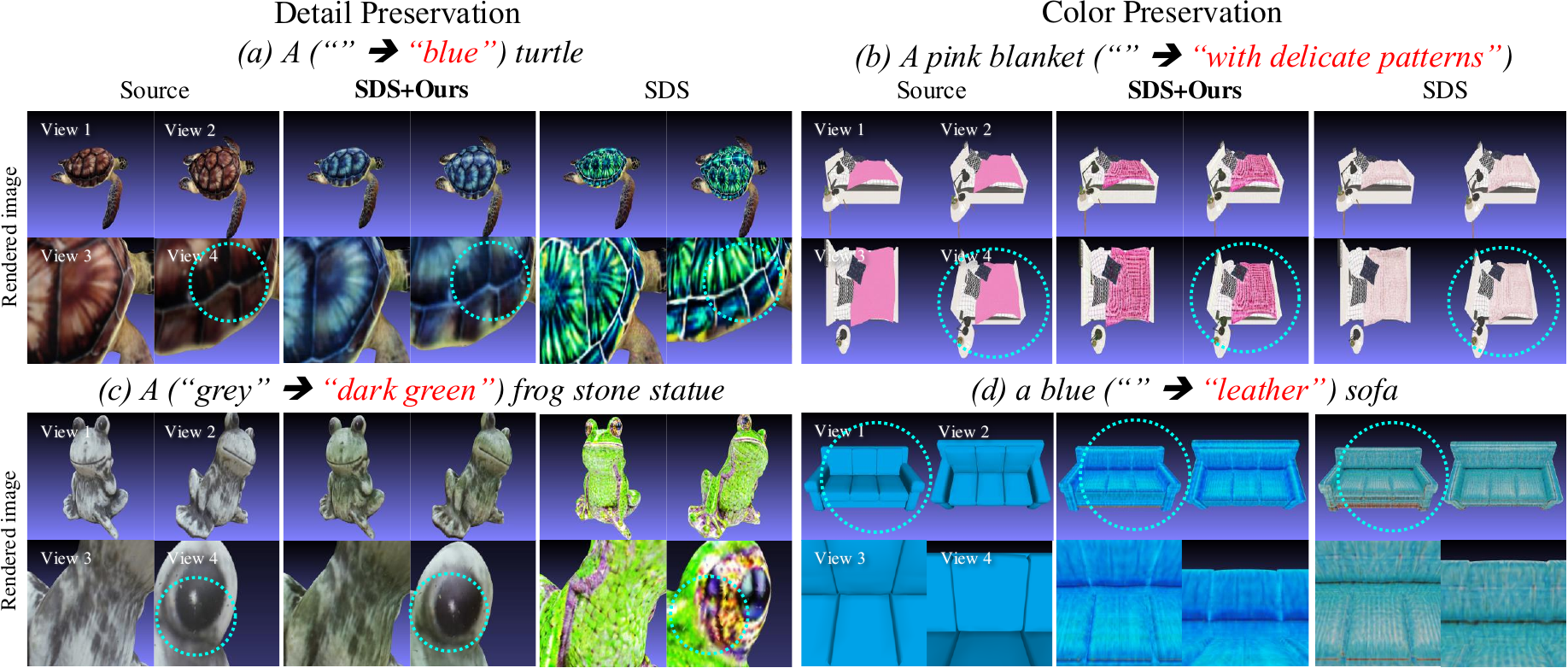} \caption{\textbf{Qualitative results.} Our frequency-aware denoising score method on 3D texture. We compare our results with the pure triplane representation. In the detail preservation cases (a) and (c), our method faithfully preserves the original texture from the source texture map. In contrast, SDS distorts the original texture and does not follow the text "blue" in the turtle case. In the color fidelity cases (b) and (d), our method retain the pink color of the quilt and the blue color of the sofa, while adding intricate patterns and texture details guided by the text. However, SDS creates patterns on the quilt but changes the main color to a much lighter pink. In the sofa case, SDS also adds texture details but shifts the main color to a greenish-blue hue whitish-red hue. Best viewed on a screen when zoomed in. }  \label{fig:qualitative:3d} \end{figure*}

\subsection{Experimental Settings}
\subsubsection{2D Image Editing}

\noindent\textbf{Baseline Methods.} To verify the effectiveness of our method, we conducted comparative experiments in 2D. Since our method is optimization-based, we primarily compare it with state-of-the-art score distillation methods, including DDS~\cite{hertz2023delta} and CDS~\cite{nam2024contrastive}, as these are the most competitive baselines. 
Moreover, we report results comparing various diffusion sampling methods, including Plug-and-Play (PnP)~\cite{tumanyan2023plug}, Prompt-to-Prompt (P2P)~\cite{hertz2022prompt}, DiffuseIT~\cite{kwon2022diffusion}, FlexiEdit~\cite{koo2024flexiedit}, and FreeDiff~\cite{wu2025freediff}. We also report on DreamSampler~\cite{kim2025dreamsampler}, which unifies diffusion sampling and score distillation.  

\noindent\textbf{Selecting Frequency Components.} As mentioned previously, we showcase our method's capability in two primary editing scenarios: color editing and detail editing, where we optimize the low-frequency and high-frequency components\footnote{We provide additional experimental results beyond these two primary settings, such as combining ours with FlexiEdit~\cite{koo2024flexiedit} to offer non-rigid editing capabilities, in the supplementary materials.}. Since selecting specific frequency components for optimization is a unique advantage of our method—one not supported by baseline methods, we compare with the baseline methods using their closest settings, such as the style modification in FreeDiff~\cite{wu2025freediff}, when applicable. 

\noindent\textbf{Evaluation Protocol.} For quantitative results, we sample 200 cat images from the LAION-COCO dataset~\cite{schuhmann2022laion} and use the editing methods to transform them into dogs, pigs and cows. 
We use the Structural Similarity Index Measure (SSIM) to assess fidelity and LPIPS~\cite{zhang2018unreasonable} to quantify the preservation of high-frequency details. 
We also measure the CLIP score~\cite{hessel2021clipscore} to evaluate alignment between edits and target text descriptions.

\subsubsection{3D Texture Editing}

We extend our method directly from 2D to 3D texture editing to demonstrate the benefits of frequency disentanglement offered by our approach. 
Firstly, since diffusion sampling methods~\cite{hertz2022prompt} cannot be applied to 3D texture editing at inference time, we experiment with score distillation methods only.
Secondly, we observe that some score distillation methods such as DDS and CDS do not perform well in 3D texture editing, we limit our comparison to SDS, to test the effect of our frequency-aware denoising score.
Similar to the 2D tasks, we examine performance differences in both low-frequency color editing and high-frequency detail editing tasks.

\begin{figure*}[!t] \centering \includegraphics[width=1.0\textwidth, page=6]{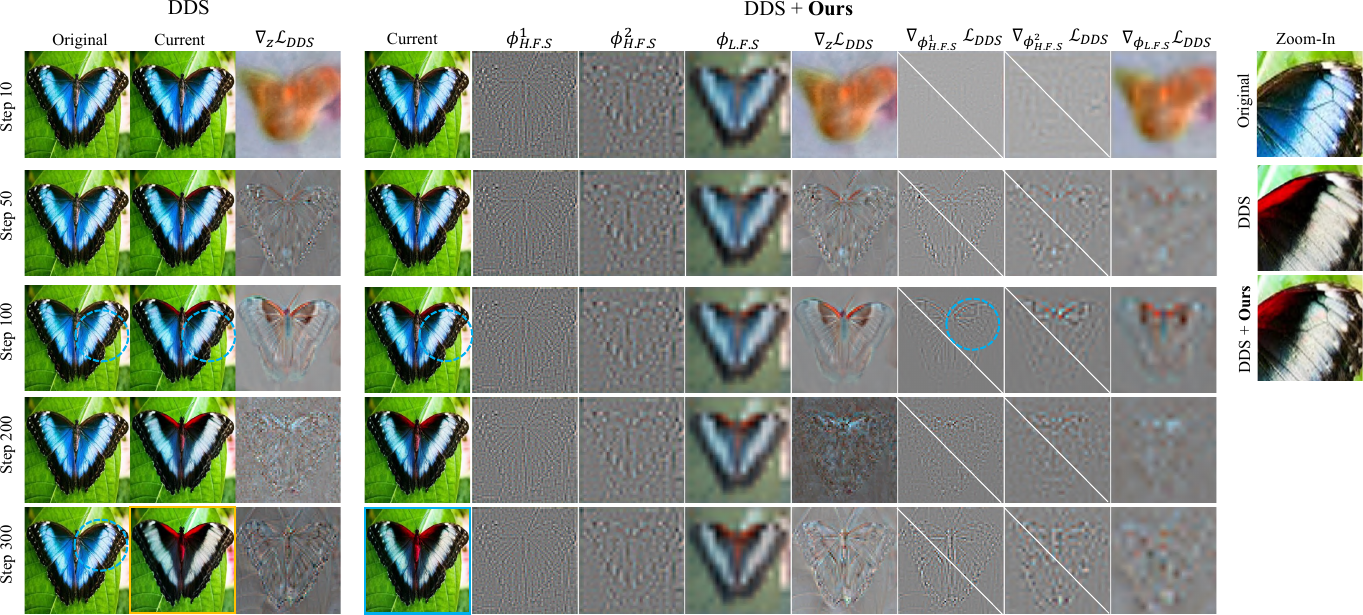} \caption{\textbf{Gradient visualization during optimization.} Edit with ``A photo of a (blue $\rightarrow$ red) butterfly". We visualize the frequency-decomposed representation $\phi$ (with three subbands, $\phi_\text{H.F.S.}^1$, $\phi_\text{H.F.S.}^2$ and $\phi_\text{L.F.S.}$), the gradients at the latent level $\nabla _z \mathcal L_{\text{DDS}}$, as well as gradients at subband level $\nabla _\phi \mathcal L_{\text{DDS}}$ in multiple wavelet subbands. As shown, although the optimization points towards the target, the gradients at each step exhibit randomness ($\nabla _z \mathcal L_{\text{DDS}}$ in step 50 and step 100). This is due to the noise added during the sampling procedure and tends to smooth out or distort detail patterns. With our method, we apply a stop-gradient operator to preserve these details (right). Note that we add a diagonal line to indicate the subband gradients where the stop-gradient operator is applied. Best viewed on a screen when zoomed in.} \label{fig:gradient-visualization} \end{figure*}

\subsection{2D Image Editing Results}
\label{sec:2d_image_editing}
We present qualitative results in Fig.~\ref{fig:qualitative:2d} with additional results provided in the supplementary material. In the detail preservation case (upper part), none of the baseline methods are able to faithfully preserve the geometry of the stone lion’s face and mouth or the chicken’s eye and beak. Among score distillation methods, CDS removes the lion’s teeth and changes the size of the chicken’s eye. DDS, on which our method is based, also removes the lion’s teeth and alters the size of its eye. Diffusion sampling methods often significantly change the structure and may fail to generate meaningful results, e.g., DreamSampler~\cite{kim2025dreamsampler} and DiffuseIT~\cite{kwon2022diffusion}. 

In the color fidelity case (lower part), DDS and CDS often disrupt color consistency by adding white to the gray cat or turning the stone yellowish. Diffusion sampling methods frequently fail to preserve color, as seen in the cat-to-fox and shark cases with PnP. In contrast, since our method explicitly disentangles high-frequency and low-frequency components and optimizes the selected frequency band, our method achieves better color consistency. 

For the quantitative evaluation, we report the results against the most competitive baselines, CDS and DDS, in Tab.~\ref{table:2D_editing_results}. Our method achieves the same CLIP score as CDS, indicating that our method performs as well as CDS in matching the text guidance. Additionally, our method achieves slightly better scores on LPIPS and SSIM, indicating improved structural preservation and consistency.

\begin{table}[h!] 
\centering
\resizebox{0.85\columnwidth}{!}{
\begin{tabular}{l l c c c} 
\toprule 
\multicolumn{2}{c}{Method} & CLIP$\uparrow$ & LPIPS $\downarrow$ & SSIM $\uparrow$ \\ 
\midrule
\multirow{2}{*}{Diffusion Sampling} & NMG~\cite{cho2024noise} & \textbf{23.68} & \underline{0.182} & 0.738 \\
 & Direct Inversion~\cite{ju2023direct} & 23.61 & 0.284 & 0.676 \\ 
\midrule
\multirow{3}{*}{Score Distillation} & DDS~\cite{hertz2023delta} & 23.64 & 0.297 & 0.736 \\ 
 & CDS~\cite{nam2024contrastive} & \underline{23.65} & \underline{0.136} & \underline{0.818} \\ 
 & Ours & \underline{23.65} & \textbf{0.129} & \textbf{0.819} \\  
\bottomrule
\end{tabular}
}
\caption{\textbf{Quantitative results on 2D image editing}. Our method achieves comparable text alignment (CLIP score) with CDS while outperforming DDS and CDS in both SSIM and LPIPS.} 
\label{table:2D_editing_results} 
\end{table}

In Tab.~\ref{table:user-study}, we present the user study results. We surveyed 24 users using six source images, each accompanied by two edited versions. Half of these images focused on detail preservation, and the other half on color fidelity. Each user was shown a set of three images: the source image, an edit labeled as option one, and another edit labeled as option two. We asked each user to select the preferred option between the two edits that better ``reflects the specified text instructions, while ensuring that local detail patterns or color remain unchanged". According to the results, on average, more than 80\% of users prefer the edits made with our method over both CDS and DDS.

\begin{table}[h!] 
\centering
\resizebox{0.8\linewidth}{!}{ %
\begin{tabular}{lccc} 
\hline 
  & Color pres.$\uparrow$ & Detail pres. $\uparrow$ & Average\\ 
\hline 
Prefer ours over DDS & 73.8\% & 90.5\% & 82.1\% \\ 
Prefer ours over CDS & 92.9\% & 79.8\% & 86.3\% \\ 
\hline 
\end{tabular} 
}
\caption{\textbf{Percentage of users preferring ours over baselines} in two editing scenarios: color preservation and detail preservation. More users preferred our method in both cases. More than 90\% users prefer ours in detail preservation than DDS, and color preservation than CDS. On average, over 80\% of users favored the edits generated by our method over CDS and DDS.
}
\label{table:user-study} 
\end{table}

\subsection{Texture Editing Results}
\label{sec:texture_editing}
Similar to the 2D case, our method enables the preservation of texture details and color fidelity, Fig.~\ref{fig:qualitative:3d}.
In detail preservation, our method can change the low-frequency components of the turtle and frog, while preserving the detailed textures. In contrast, the baseline SDS changes the shell texture and shifts the frog's eyes. In the color preservation setting, ours adds high-frequency texture to the quilt and adds leather texture to the sofa, while maintaining color consistency, whereas the baseline SDS alters the color.

\section{Discussions}
\label{sec:discussion}
\subsection{Gradient Visualization in Score Distillation}
\label{sec:gradient_visualization}
In Fig.~\ref{fig:gradient-visualization}, we visualize the gradient $\nabla _z \mathcal L_{\text{DDS}}$ in DDS. It is noteworthy that although the gradient generally aligns with the text guidance (e.g., turning the blue butterfly to red), the inherent randomness of score distillation causes inconsistencies in $\nabla _z \mathcal L_{\text{DDS}}$ across multiple steps. Consequently, this noise alters the detailed structure on the wing. 
By applying our method with a stop-gradient operation to specific frequency bands, we achieve the desired editing while preserving local details (right). We provide gradient visualization for high frequency editing in supplementary material.

\begin{figure*}[t] \centering \includegraphics[width=1.0\textwidth, page=7]{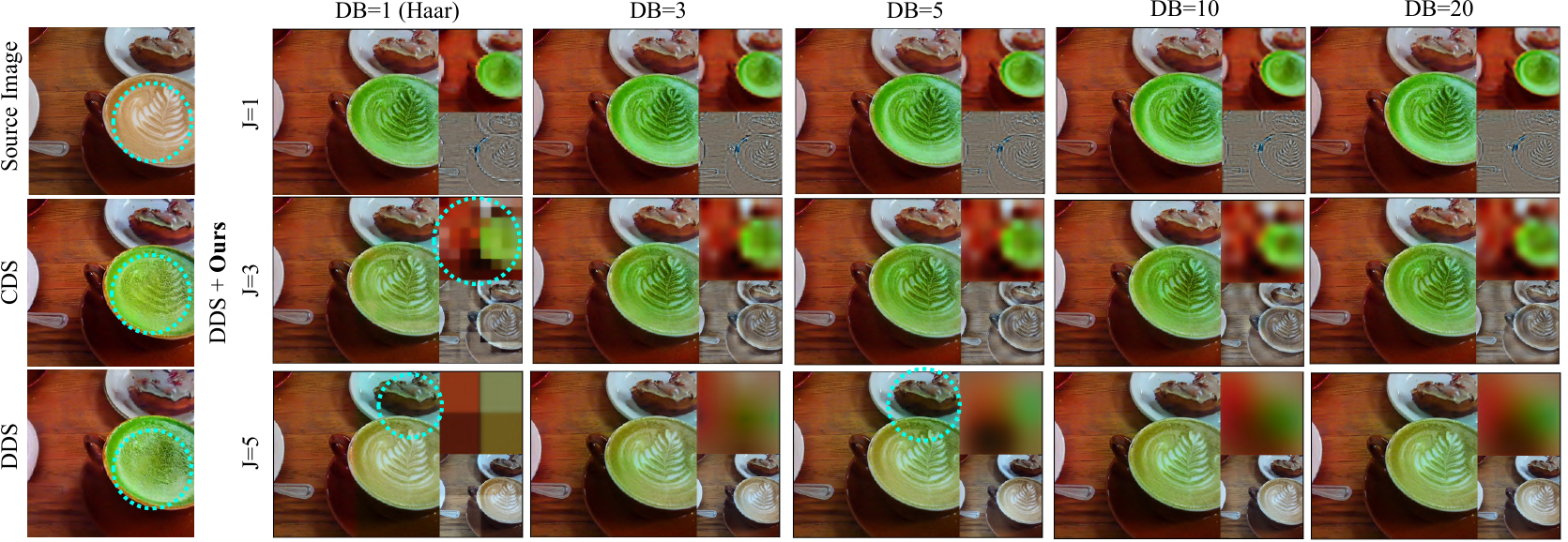} \caption{\textbf{Comparing different wavelets parameters}. Edit with ``A cup of (coffee $\rightarrow$ matcha)". Rows: Daubechies wavelets with an increasing number of vanishing moments. Columns: levels in multi-level decomposition. Larger Daubechies wavelet indices result in smoother transformations, in contrast to the mosaic-like pattern created by the Haar Wavelet. Larger J values
lead to lower-resolution low frequency components, making it difficult to model the round shape of the matcha cup at J=5. Best viewed on a screen when zoomed in.} \label{fig:db} \end{figure*}

\begin{figure}[!t]
    \centering
    \includegraphics[width=0.5\textwidth,page=8]{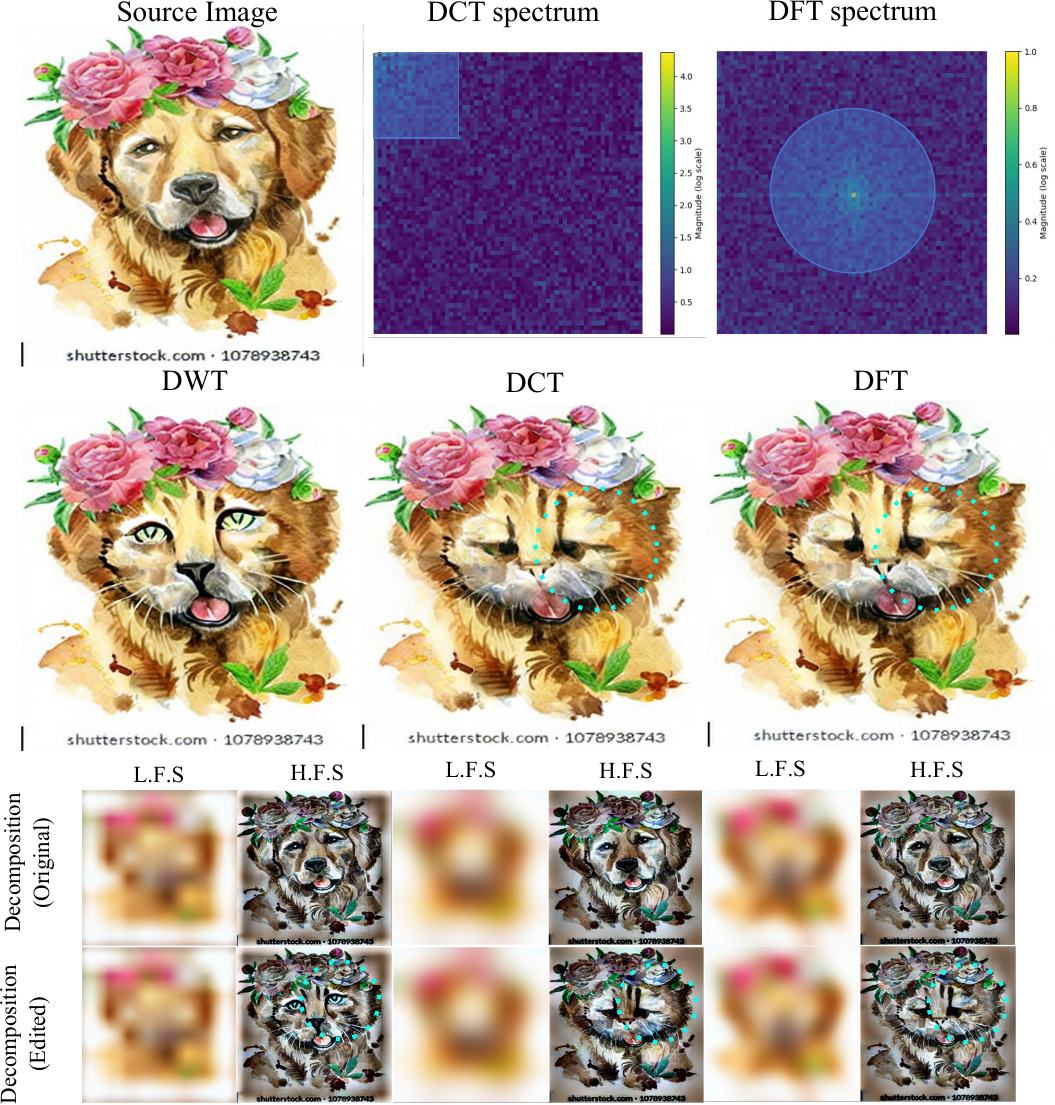}
    \caption{\textbf{Comparison of different frequency decomposition methods}. Edit with ``A drawing of a (dog $\rightarrow$ cat). Both DCT and DFT lack spatial dimension in decomposition, causing artifacts like the cat's eyes. In the spectrum, the blue areas represent the mask schematic used in this experiment. For both DFT and DCT, we applied mask to the low-frequency regions to limit parameter updates. Best viewed on a screen when zoomed in.} 
\label{fig:DCT:example}
\end{figure}

\subsection{Choosing Among the Wavelet Family}
Different wavelets have distinct filters, which enable them to extract different types of features.
In this experiment, we analyze the two attributes that most significantly impact the result: smoothness and decomposition level in the Daubechies wavelet family~\cite{daubechies1993ten}, Fig.~\ref{fig:db}. We use the instruction to turn ``a coffee latte art" into ``a matcha latte art" while editing the low-frequency component. 
Due to the halving nature of wavelets, a higher J value results in a lower resolution for the low-frequency subband. 
We observe reduced localization ability at high J levels (e.g., J=5), causing color to spill over into adjacent regions. 
Regarding smoothness, when DB=1, Daubechies wavelet reduces to the Haar wavelet~\cite{haar1911theorie}, featuring sharp edges, while a larger index produces a smoother result. 
Generally, a medium setting (e.g., J=3 and DB=3) yields decent outputs.

\subsection{Comparing Other Frequency Methods}
\label{fig:DCT}
Wavelets provide distinct advantages over other frequency decomposition methods, such as superior spatial localization and multiresolution decomposition capabilities. 
We compare wavelets with DFT and DCT in Fig.~\ref{fig:DCT:example}. Regarding the DFT and DCT parameters, we select a frequency threshold that achieves high-low frequency separation similar to ours (original row). 
Although all three decomposition methods enable frequency separation, already improving upon DDS, we find that the superior localization capability of wavelets makes them a better choice in certain cases.

\subsection{Limitations}

While our selective frequency optimization improves color fidelity and detail preservation for precise editing tasks, it inherits the limitations of diffusion models and score distillation, such as challenges in accurately interpreting prompts and requiring a lengthy sampling process

\section{Conclusion}

In this paper, we identify the issue of indiscriminate optimization across all frequency subbands in score distillation text-guided image editing. 
To address this, we propose a frequency-aware denoising score that utilizes the discrete wavelet transform to selectively optimize specific frequency subbands. This approach achieves improved preservation of high-frequency details and enhances color consistency.
We hope our proposed method will pave the way for fine-grained, controllable image and texture editing.

\newpage
\section{Acknowledgment} 
Zicong Jiang is currently affiliated with Chalmers University of Technology. He made significant contributions to the code implementations and experimental validations during his master's thesis work at IVRL.
The authors extend their gratitude to Quentin Bammey, Raphaël Achddou, Dongqing Wang, and Martin Everaert for their thorough proofreading and valuable feedback, as well as to the reviewers for their constructive suggestions.
{
    \small
    \bibliographystyle{ieeenat_fullname}
    \bibliography{main}

\begin{thebibliography}{51}
\providecommand{\natexlab}[1]{#1}
\providecommand{\url}[1]{\texttt{#1}}
\expandafter\ifx\csname urlstyle\endcsname\relax
  \providecommand{\doi}[1]{doi: #1}\else
  \providecommand{\doi}{doi: \begingroup \urlstyle{rm}\Url}\fi

\bibitem[Berzak et~al.(2016)Berzak, Barbu, Harari, Katz, and Ullman]{berzak2016you}
Yevgeni Berzak, Andrei Barbu, Daniel Harari, Boris Katz, and Shimon Ullman.
\newblock Do you see what i mean? visual resolution of linguistic ambiguities.
\newblock \emph{arXiv preprint arXiv:1603.08079}, 2016.

\bibitem[Brooks et~al.(2023)Brooks, Holynski, and Efros]{brooks2023instructpix2pix}
Tim Brooks, Aleksander Holynski, and Alexei~A Efros.
\newblock Instructpix2pix: Learning to follow image editing instructions.
\newblock In \emph{CVPR}, pages 18392--18402, 2023.

\bibitem[Cai et~al.(2021)Cai, Zhang, Huang, Geng, Li, and Huang]{cai2021frequency}
Mu Cai, Hong Zhang, Huijuan Huang, Qichuan Geng, Yixuan Li, and Gao Huang.
\newblock Frequency domain image translation: More photo-realistic, better identity-preserving.
\newblock In \emph{CVPR}, pages 13930--13940, 2021.

\bibitem[Cao et~al.(2023)Cao, Wang, Qi, Shan, Qie, and Zheng]{cao2023masactrl}
Mingdeng Cao, Xintao Wang, Zhongang Qi, Ying Shan, Xiaohu Qie, and Yinqiang Zheng.
\newblock Masactrl: Tuning-free mutual self-attention control for consistent image synthesis and editing.
\newblock In \emph{ICCV}, pages 22560--22570, 2023.

\bibitem[Chan et~al.(2022{\natexlab{a}})Chan, Lin, Chan, Nagano, Pan, De~Mello, Gallo, Guibas, Tremblay, Khamis, et~al.]{Chan2021}
Eric~R Chan, Connor~Z Lin, Matthew~A Chan, Koki Nagano, Boxiao Pan, Shalini De~Mello, Orazio Gallo, Leonidas~J Guibas, Jonathan Tremblay, Sameh Khamis, et~al.
\newblock Efficient geometry-aware 3d generative adversarial networks.
\newblock In \emph{CVPR}, pages 16123--16133, 2022{\natexlab{a}}.

\bibitem[Chan et~al.(2022{\natexlab{b}})Chan, Lin, Chan, Nagano, Pan, De~Mello, Gallo, Guibas, Tremblay, Khamis, et~al.]{chan2022efficient}
Eric~R Chan, Connor~Z Lin, Matthew~A Chan, Koki Nagano, Boxiao Pan, Shalini De~Mello, Orazio Gallo, Leonidas~J Guibas, Jonathan Tremblay, Sameh Khamis, et~al.
\newblock Efficient geometry-aware 3d generative adversarial networks.
\newblock In \emph{CVPR}, pages 16123--16133, 2022{\natexlab{b}}.

\bibitem[Cho et~al.(2024)Cho, Lee, Kim, Oh, and Jeong]{cho2024noise}
Hansam Cho, Jonghyun Lee, Seoung~Bum Kim, Tae-Hyun Oh, and Yonghyun Jeong.
\newblock Noise map guidance: Inversion with spatial context for real image editing.
\newblock \emph{arXiv preprint arXiv:2402.04625}, 2024.

\bibitem[Corneanu et~al.(2024)Corneanu, Gadde, and Martinez]{corneanu2024latentpaint}
Ciprian Corneanu, Raghudeep Gadde, and Aleix~M Martinez.
\newblock Latentpaint: Image inpainting in latent space with diffusion models.
\newblock In \emph{WACV}, pages 4334--4343, 2024.

\bibitem[Cotter(2020)]{cotter2020uses}
Fergal Cotter.
\newblock \emph{Uses of complex wavelets in deep convolutional neural networks}.
\newblock PhD thesis, 2020.

\bibitem[Couairon et~al.(2022)Couairon, Verbeek, Schwenk, and Cord]{couairon2022diffedit}
Guillaume Couairon, Jakob Verbeek, Holger Schwenk, and Matthieu Cord.
\newblock Diffedit: Diffusion-based semantic image editing with mask guidance.
\newblock \emph{arXiv preprint arXiv:2210.11427}, 2022.

\bibitem[Daubechies(1992)]{daubechies1993ten}
Ingrid Daubechies.
\newblock \emph{Ten lectures on wavelets}.
\newblock SIAM, 1992.

\bibitem[Drori et~al.(2003)Drori, Cohen-Or, and Yeshurun]{drori2003fragment}
Iddo Drori, Daniel Cohen-Or, and Hezy Yeshurun.
\newblock Fragment-based image completion.
\newblock \emph{ACM Trans. Graph.}, 22\penalty0 (3):\penalty0 303–312, 2003.

\bibitem[Everaert et~al.(2024)Everaert, Fitsios, Bocchio, Arpa, S{\"u}sstrunk, and Achanta]{everaert2024exploiting}
Martin~Nicolas Everaert, Athanasios Fitsios, Marco Bocchio, Sami Arpa, Sabine S{\"u}sstrunk, and Radhakrishna Achanta.
\newblock Exploiting the signal-leak bias in diffusion models.
\newblock In \emph{WACV}, pages 4025--4034, 2024.

\bibitem[Gao et~al.(2024)Gao, Xu, Zhao, and Liu]{gao2024frequency}
Xiang Gao, Zhengbo Xu, Junhan Zhao, and Jiaying Liu.
\newblock Frequency-controlled diffusion model for versatile text-guided image-to-image translation.
\newblock In \emph{AAAI}, pages 1824--1832, 2024.

\bibitem[Haar(1911)]{haar1911theorie}
Alfred Haar.
\newblock Zur theorie der orthogonalen funktionensysteme.
\newblock \emph{Mathematische Annalen}, 71\penalty0 (1):\penalty0 38--53, 1911.

\bibitem[Hertz et~al.(2022)Hertz, Mokady, Tenenbaum, Aberman, Pritch, and Cohen-Or]{hertz2022prompt}
Amir Hertz, Ron Mokady, Jay Tenenbaum, Kfir Aberman, Yael Pritch, and Daniel Cohen-Or.
\newblock Prompt-to-prompt image editing with cross attention control.
\newblock \emph{arXiv preprint arXiv:2208.01626}, 2022.

\bibitem[Hertz et~al.(2023)Hertz, Aberman, and Cohen-Or]{hertz2023delta}
Amir Hertz, Kfir Aberman, and Daniel Cohen-Or.
\newblock Delta denoising score.
\newblock In \emph{ICCV}, pages 2328--2337, 2023.

\bibitem[Hessel et~al.(2021)Hessel, Holtzman, Forbes, Bras, and Choi]{hessel2021clipscore}
Jack Hessel, Ari Holtzman, Maxwell Forbes, Ronan~Le Bras, and Yejin Choi.
\newblock {CLIPScore:} a reference-free evaluation metric for image captioning.
\newblock In \emph{EMNLP}, 2021.

\bibitem[Ju et~al.(2023)Ju, Zeng, Bian, Liu, and Xu]{ju2023direct}
Xuan Ju, Ailing Zeng, Yuxuan Bian, Shaoteng Liu, and Qiang Xu.
\newblock Direct inversion: Boosting diffusion-based editing with 3 lines of code.
\newblock \emph{arXiv preprint arXiv:2310.01506}, 2023.

\bibitem[Karras et~al.(2022)Karras, Aittala, Aila, and Laine]{karras2022elucidating}
Tero Karras, Miika Aittala, Timo Aila, and Samuli Laine.
\newblock Elucidating the design space of diffusion-based generative models.
\newblock \emph{NeurIPS}, 35:\penalty0 26565--26577, 2022.

\bibitem[Kawar et~al.(2023)Kawar, Zada, Lang, Tov, Chang, Dekel, Mosseri, and Irani]{kawar2023imagic}
Bahjat Kawar, Shiran Zada, Oran Lang, Omer Tov, Huiwen Chang, Tali Dekel, Inbar Mosseri, and Michal Irani.
\newblock Imagic: Text-based real image editing with diffusion models.
\newblock In \emph{CVPR}, pages 6007--6017, 2023.

\bibitem[Ke et~al.(2024)Ke, Obukhov, Huang, Metzger, Daudt, and Schindler]{ke2024repurposing}
Bingxin Ke, Anton Obukhov, Shengyu Huang, Nando Metzger, Rodrigo~Caye Daudt, and Konrad Schindler.
\newblock Repurposing diffusion-based image generators for monocular depth estimation.
\newblock In \emph{CVPR}, pages 9492--9502, 2024.

\bibitem[Kim et~al.(2025)Kim, Park, and Ye]{kim2025dreamsampler}
Jeongsol Kim, Geon~Yeong Park, and Jong~Chul Ye.
\newblock Dreamsampler: Unifying diffusion sampling and score distillation for image manipulation.
\newblock In \emph{ECCV}, pages 398--414. Springer, 2025.

\bibitem[Kingma et~al.(2013)Kingma, Welling, et~al.]{kingma2013auto}
Diederik~P Kingma, Max Welling, et~al.
\newblock Auto-encoding variational bayes, 2013.

\bibitem[Koo et~al.(2024{\natexlab{a}})Koo, Yoon, Hong, and Yoo]{koo2024flexiedit}
Gwanhyeong Koo, Sunjae Yoon, Ji~Woo Hong, and Chang~D Yoo.
\newblock Flexiedit: Frequency-aware latent refinement for enhanced non-rigid editing.
\newblock In \emph{ECCV}, pages 363--379. Springer, 2024{\natexlab{a}}.

\bibitem[Koo et~al.(2024{\natexlab{b}})Koo, Park, and Sung]{koo2024posterior}
Juil Koo, Chanho Park, and Minhyuk Sung.
\newblock Posterior distillation sampling.
\newblock In \emph{Proceedings of the IEEE/CVF Conference on Computer Vision and Pattern Recognition}, pages 13352--13361, 2024{\natexlab{b}}.

\bibitem[Krishnamoorthi and Malarchelvi(2009)]{krishnamoorthi2009image}
R Krishnamoorthi and Sheba~Kezia Malarchelvi.
\newblock Image adaptive watermarking with visual model in orthogonal polynomials based transformation domain.
\newblock \emph{International Journal of Signal Processing}, 5\penalty0 (2):\penalty0 146--153, 2009.

\bibitem[Kwon and Ye(2022)]{kwon2022diffusion}
Gihyun Kwon and Jong~Chul Ye.
\newblock Diffusion-based image translation using disentangled style and content representation.
\newblock \emph{arXiv preprint arXiv:2209.15264}, 2022.

\bibitem[Mehrabi et~al.(2023)Mehrabi, Goyal, Verma, Dhamala, Kumar, Hu, Chang, Zemel, Galstyan, and Gupta]{mehrabi2023resolving}
Ninareh Mehrabi, Palash Goyal, Apurv Verma, Jwala Dhamala, Varun Kumar, Qian Hu, Kai-Wei Chang, Richard Zemel, Aram Galstyan, and Rahul Gupta.
\newblock Resolving ambiguities in text-to-image generative models.
\newblock In \emph{Proceedings of the 61st Annual Meeting of the Association for Computational Linguistics (Volume 1: Long Papers)}, pages 14367--14388, 2023.

\bibitem[Meng et~al.(2021)Meng, Song, Song, Wu, Zhu, and Ermon]{meng2021sdedit}
Chenlin Meng, Yang Song, Jiaming Song, Jiajun Wu, Jun-Yan Zhu, and Stefano Ermon.
\newblock Sdedit: Image synthesis and editing with stochastic differential equations.
\newblock \emph{arXiv preprint arXiv:2108.01073}, 2021.

\bibitem[Miyake et~al.(2023)Miyake, Iohara, Saito, and Tanaka]{miyake2023negative}
Daiki Miyake, Akihiro Iohara, Yu Saito, and Toshiyuki Tanaka.
\newblock Negative-prompt inversion: Fast image inversion for editing with text-guided diffusion models.
\newblock \emph{arXiv preprint arXiv:2305.16807}, 2023.

\bibitem[Mokady et~al.(2023)Mokady, Hertz, Aberman, Pritch, and Cohen-Or]{mokady2023null}
Ron Mokady, Amir Hertz, Kfir Aberman, Yael Pritch, and Daniel Cohen-Or.
\newblock Null-text inversion for editing real images using guided diffusion models.
\newblock In \emph{CVPR}, pages 6038--6047, 2023.

\bibitem[Moser et~al.(2024)Moser, Shanbhag, Raue, Frolov, Palacio, and Dengel]{moser2024diffusion}
Brian~B Moser, Arundhati~S Shanbhag, Federico Raue, Stanislav Frolov, Sebastian Palacio, and Andreas Dengel.
\newblock Diffusion models, image super-resolution, and everything: A survey.
\newblock \emph{IEEE Transactions on Neural Networks and Learning Systems}, 2024.

\bibitem[Nam et~al.(2024)Nam, Kwon, Park, and Ye]{nam2024contrastive}
Hyelin Nam, Gihyun Kwon, Geon~Yeong Park, and Jong~Chul Ye.
\newblock Contrastive denoising score for text-guided latent diffusion image editing.
\newblock In \emph{CVPR}, pages 9192--9201, 2024.

\bibitem[Pan et~al.(2024)Pan, Zhang, Chen, Zhou, Ke, S{\"u}sstrunk, and Salzmann]{pan2024coherent}
Lingzhi Pan, Tong Zhang, Bingyuan Chen, Qi Zhou, Wei Ke, Sabine S{\"u}sstrunk, and Mathieu Salzmann.
\newblock Coherent and multi-modality image inpainting via latent space optimization.
\newblock \emph{arXiv preprint arXiv:2407.08019}, 2024.

\bibitem[Paszke et~al.(2019)Paszke, Gross, Massa, Lerer, Bradbury, Chanan, Killeen, Lin, Gimelshein, Antiga, Desmaison, Kopf, Yang, DeVito, Raison, Tejani, Chilamkurthy, Steiner, Fang, Bai, and Chintala]{NEURIPS2019_9015}
Adam Paszke, Sam Gross, Francisco Massa, Adam Lerer, James Bradbury, Gregory Chanan, Trevor Killeen, Zeming Lin, Natalia Gimelshein, Luca Antiga, Alban Desmaison, Andreas Kopf, Edward Yang, Zachary DeVito, Martin Raison, Alykhan Tejani, Sasank Chilamkurthy, Benoit Steiner, Lu Fang, Junjie Bai, and Soumith Chintala.
\newblock Pytorch: An imperative style, high-performance deep learning library.
\newblock In \emph{Advances in Neural Information Processing Systems 32}, pages 8024--8035. Curran Associates, Inc., 2019.

\bibitem[Poole et~al.(2022)Poole, Jain, Barron, and Mildenhall]{poole2022dreamfusion}
Ben Poole, Ajay Jain, Jonathan~T Barron, and Ben Mildenhall.
\newblock Dreamfusion: Text-to-3d using 2d diffusion.
\newblock \emph{arXiv preprint arXiv:2209.14988}, 2022.

\bibitem[Prasad et~al.(2023)Prasad, Stengel-Eskin, and Bansal]{prasad2023rephrase}
Archiki Prasad, Elias Stengel-Eskin, and Mohit Bansal.
\newblock Rephrase, augment, reason: Visual grounding of questions for vision-language models.
\newblock \emph{arXiv preprint arXiv:2310.05861}, 2023.

\bibitem[Ramesh et~al.(2022)Ramesh, Dhariwal, Nichol, Chu, and Chen]{ramesh2022hierarchical}
Aditya Ramesh, Prafulla Dhariwal, Alex Nichol, Casey Chu, and Mark Chen.
\newblock Hierarchical text-conditional image generation with clip latents.
\newblock \emph{arXiv preprint arXiv:2204.06125}, 1\penalty0 (2):\penalty0 3, 2022.

\bibitem[Rombach et~al.(2022)Rombach, Blattmann, Lorenz, Esser, and Ommer]{rombach2021highresolution}
Robin Rombach, Andreas Blattmann, Dominik Lorenz, Patrick Esser, and Bj{\"o}rn Ommer.
\newblock High-resolution image synthesis with latent diffusion models.
\newblock In \emph{CVPR}, pages 10684--10695, 2022.

\bibitem[Saharia et~al.(2022)Saharia, Chan, Saxena, Li, Whang, Denton, Ghasemipour, Gontijo~Lopes, Karagol~Ayan, Salimans, et~al.]{saharia2022photorealistic}
Chitwan Saharia, William Chan, Saurabh Saxena, Lala Li, Jay Whang, Emily~L Denton, Kamyar Ghasemipour, Raphael Gontijo~Lopes, Burcu Karagol~Ayan, Tim Salimans, et~al.
\newblock Photorealistic text-to-image diffusion models with deep language understanding.
\newblock \emph{NeurIPS}, 35:\penalty0 36479--36494, 2022.

\bibitem[Schuhmann et~al.(2022)Schuhmann, Beaumont, Vencu, Gordon, Wightman, Cherti, Coombes, Katta, Mullis, Wortsman, et~al.]{schuhmann2022laion}
Christoph Schuhmann, Romain Beaumont, Richard Vencu, Cade Gordon, Ross Wightman, Mehdi Cherti, Theo Coombes, Aarush Katta, Clayton Mullis, Mitchell Wortsman, et~al.
\newblock Laion-5b: An open large-scale dataset for training next generation image-text models.
\newblock \emph{NeurIPS}, 35:\penalty0 25278--25294, 2022.

\bibitem[Tumanyan et~al.(2023)Tumanyan, Geyer, Bagon, and Dekel]{tumanyan2023plug}
Narek Tumanyan, Michal Geyer, Shai Bagon, and Tali Dekel.
\newblock Plug-and-play diffusion features for text-driven image-to-image translation.
\newblock In \emph{CVPR}, pages 1921--1930, 2023.

\bibitem[Wang et~al.(2024)Wang, Yue, Zhou, Chan, and Loy]{wang2024exploiting}
Jianyi Wang, Zongsheng Yue, Shangchen Zhou, Kelvin~CK Chan, and Chen~Change Loy.
\newblock Exploiting diffusion prior for real-world image super-resolution.
\newblock \emph{IJCV}, pages 1--21, 2024.

\bibitem[Wu et~al.(2024)Wu, Fan, Qin, Gu, Zhao, and Chan]{wu2025freediff}
Wei Wu, Qingnan Fan, Shuai Qin, Hong Gu, Ruoyu Zhao, and Antoni~B Chan.
\newblock Freediff: Progressive frequency truncation for image editing with diffusion models.
\newblock In \emph{ECCV}, pages 194--209. Springer, 2024.

\bibitem[Xie et~al.(2023)Xie, Zhang, Lin, Hinz, and Zhang]{xie2023smartbrush}
Shaoan Xie, Zhifei Zhang, Zhe Lin, Tobias Hinz, and Kun Zhang.
\newblock Smartbrush: Text and shape guided object inpainting with diffusion model.
\newblock In \emph{CVPR}, pages 22428--22437, 2023.

\bibitem[Yang et~al.(2022)Yang, He, Xu, and Gao]{yang2022elegant}
Chenyu Yang, Wanrong He, Yingqing Xu, and Yang Gao.
\newblock Elegant: Exquisite and locally editable gan for makeup transfer.
\newblock In \emph{ECCV}, pages 737--754. Springer, 2022.

\bibitem[Yang et~al.(2010)Yang, Su, and Sun]{yang2010medical}
Yi Yang, Zhengwei Su, and L Sun.
\newblock Medical image enhancement algorithm based on wavelet transform.
\newblock \emph{Electronics letters}, 46\penalty0 (2):\penalty0 120--121, 2010.

\bibitem[Yu et~al.(2022)Yu, Zheng, Zhou, Huang, Xiao, and Zhao]{yu2022frequency}
Hu Yu, Naishan Zheng, Man Zhou, Jie Huang, Zeyu Xiao, and Feng Zhao.
\newblock Frequency and spatial dual guidance for image dehazing.
\newblock In \emph{ECCV}, pages 181--198. Springer, 2022.

\bibitem[Zhang et~al.(2023)Zhang, Rao, and Agrawala]{zhang2023adding}
Lvmin Zhang, Anyi Rao, and Maneesh Agrawala.
\newblock Adding conditional control to text-to-image diffusion models.
\newblock In \emph{ICCV}, pages 3836--3847, 2023.

\bibitem[Zhang et~al.(2018)Zhang, Isola, Efros, Shechtman, and Wang]{zhang2018unreasonable}
Richard Zhang, Phillip Isola, Alexei~A Efros, Eli Shechtman, and Oliver Wang.
\newblock The unreasonable effectiveness of deep features as a perceptual metric.
\newblock In \emph{CVPR}, pages 586--595, 2018.

\end{thebibliography}
}

\clearpage
\setcounter{page}{1}
\maketitlesupplementary

\begin{figure}[h] \centering \includegraphics[width=0.4\textwidth, page=1]{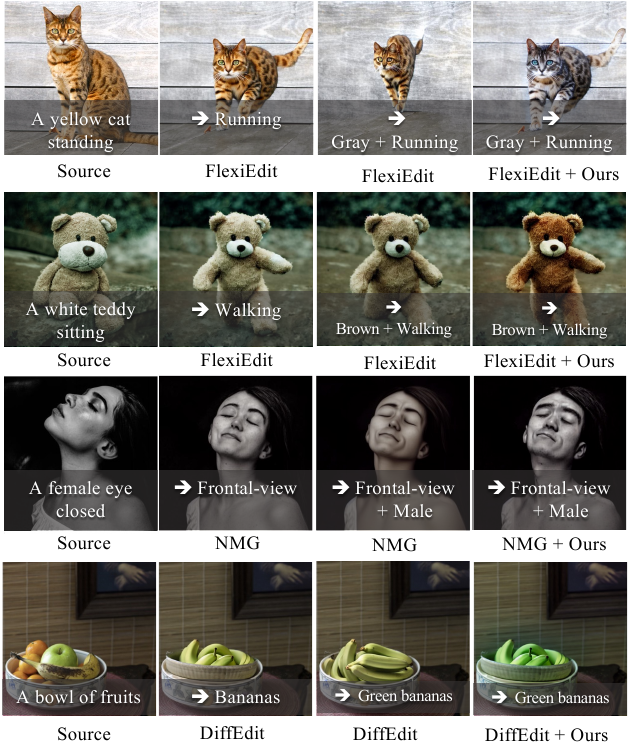} \caption{\textbf{Our method complements other editing methods.}. 
}  \label{supp:fig:complement} \end{figure}

\noindent \textbf{Combining our method with other editing methods.} As shown in Fig.~\ref{supp:fig:complement}, our method can be combined with other editing methods, i.e., FlexiEdit~\cite{koo2024flexiedit}, NMG~\cite{cho2024noise}, and DiffEdit~\cite{couairon2022diffedit}, to enable complex editing tasks.

\begin{figure}[t] \centering \includegraphics[width=0.435\textwidth, page=2]{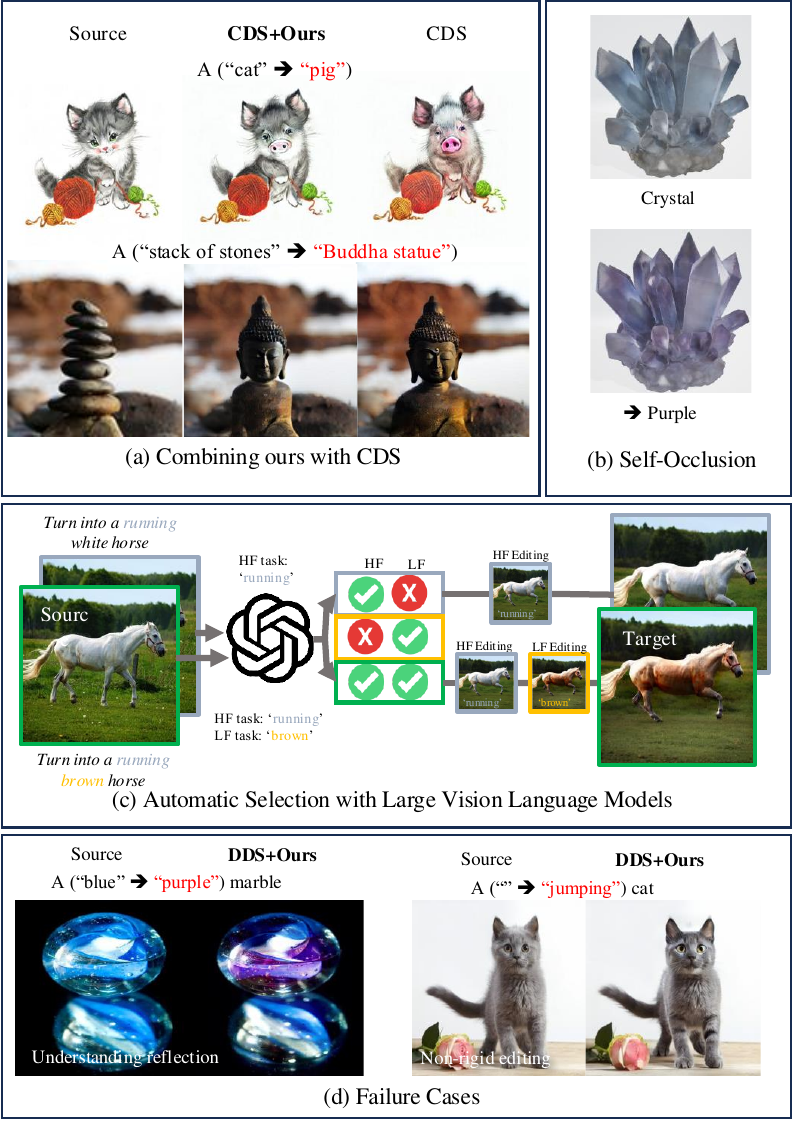} \caption{\textbf{Additional visualizations. 
}  \label{supp:fig:supp-merged}} \end{figure}

\begin{figure*}[t] \centering \includegraphics[width=1.0\textwidth, page=11]{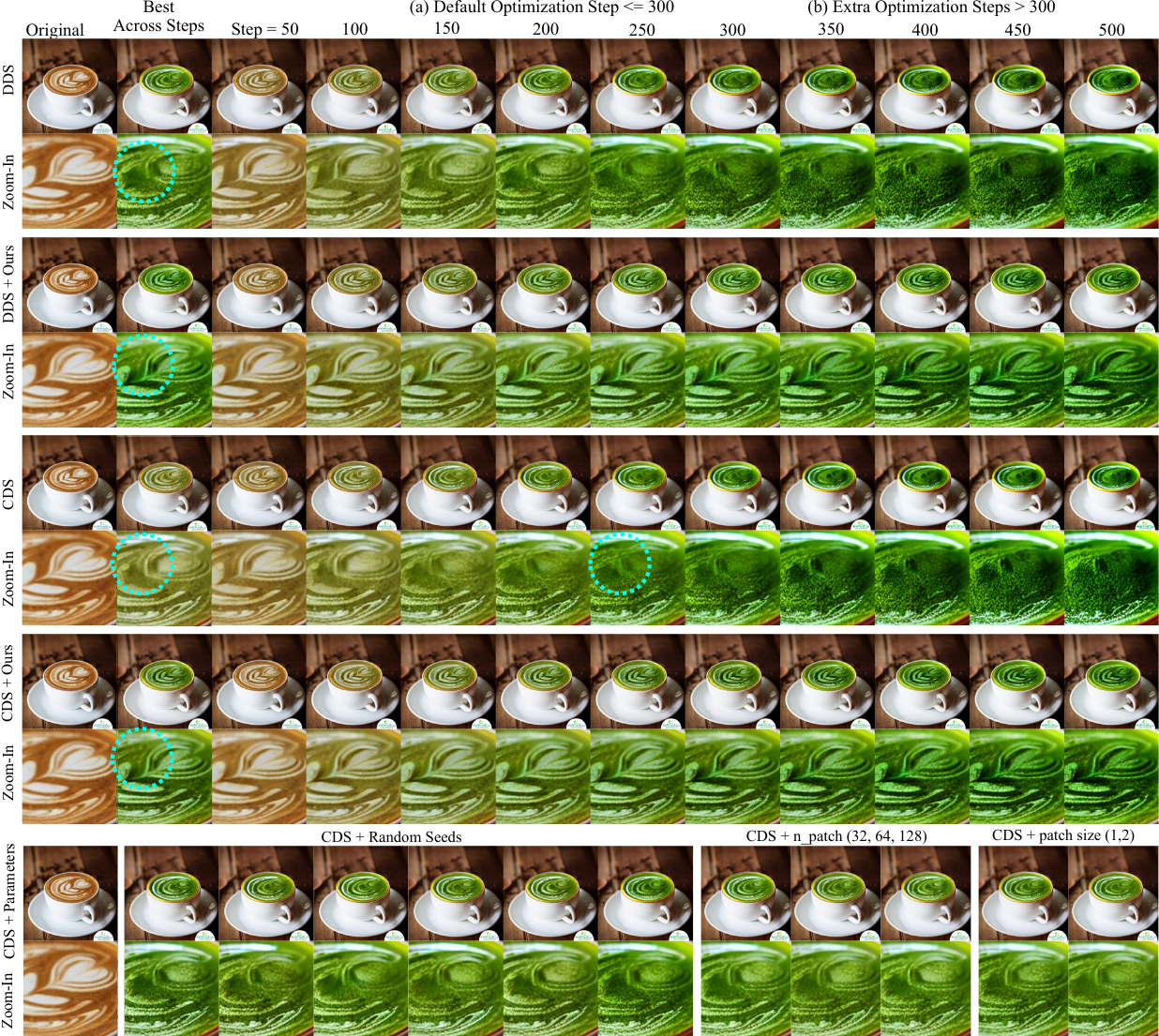} \caption{\textbf{Combining CDS with our wavelet representation.} Neither vanilla DDS nor vanilla SDS successfully preserves the intricate patterns of latte art. Notably, the distortion of details occurs simultaneously with the color change to green. By incorporating our frequency-aware representation, we achieve superior detail preservation, even when the number of steps is significantly increased, such as to 500. Additionally, we provide further results demonstrating the performance of CDS under varying hyperparameters and random seeds. (Best viewed on a screen when zoomed in)}  \label{supp:fig:cds_ours} \end{figure*}

\noindent \textbf{Combining Ours with CDS.} Our wavelet representation can also be integrated with CDS (see Fig.~\ref{supp:fig:cds_ours} and Fig.~\ref{supp:fig:supp-merged} (a)). In Fig.~\ref{supp:fig:cds_ours}, both CDS and DDS fail to preserve the detailed latte art pattern across various hyperparameters and random seeds. However, when combined with our wavelet representation, both DDS+Ours and CDS+Ours demonstrate improved texture preservation. Fig.~\ref{supp:fig:supp-merged} (a) further highlights our color consistency benefits.

\noindent \textbf{Challenging and Failure Cases.} Fig. ~\ref{supp:fig:supp-merged} (b) illustrates a challenging 3D texture editing scenario where the crystal exhibits severe self-occlusions. In this case, our method successfully changes its color despite the complexity. However, certain edits (see Fig. ~\ref{supp:fig:supp-merged} (d)) are not physically plausible, and our method encounters difficulties when handling non-rigid editing scenarios.

\noindent \textbf{Automatic Frequency Selection.} Fig. ~\ref{supp:fig:supp-merged} (c) demonstrates a pipeline that leverages a Large Vision Language Model to automatically select the appropriate frequency subband for editing based on the given prompt.

\begin{figure*}[t] \centering \includegraphics[width=1.0\textwidth, page=3]{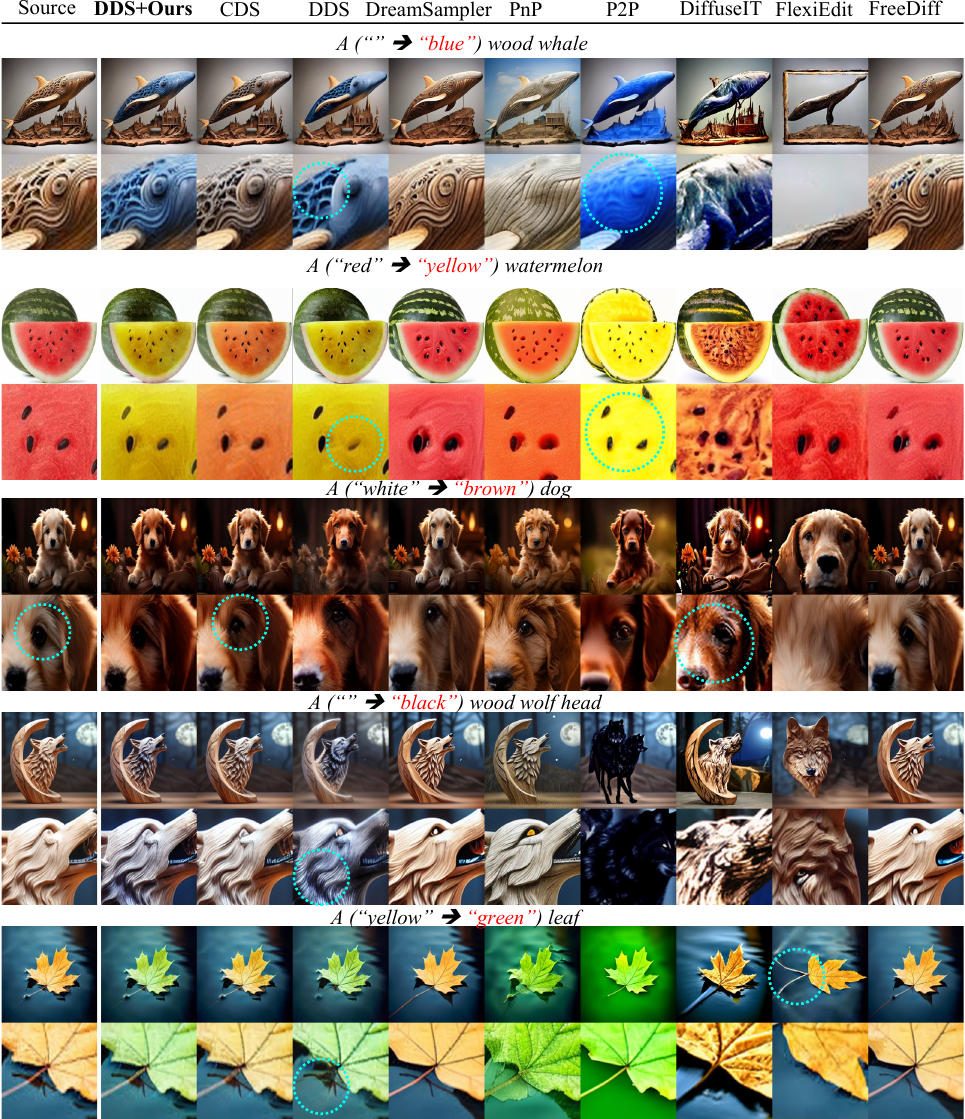} \caption{\textbf{Qualitative results on detail preservation.} Our method preserves detailed structures during editing, such as the intricate carvings in wood (rows 1 and 5) and small details like the petiole and leaf veins (last row). In contrast, CDS does not follow instructions well, and other diffusion sampling-based methods either distort the details or fail to adhere to instructions. (Best viewed on a screen when zoomed in)}  \label{supp:fig:2d-low-freq-edit1} \end{figure*}

\begin{figure*}[t] \centering \includegraphics[width=1.0\textwidth, page=4]{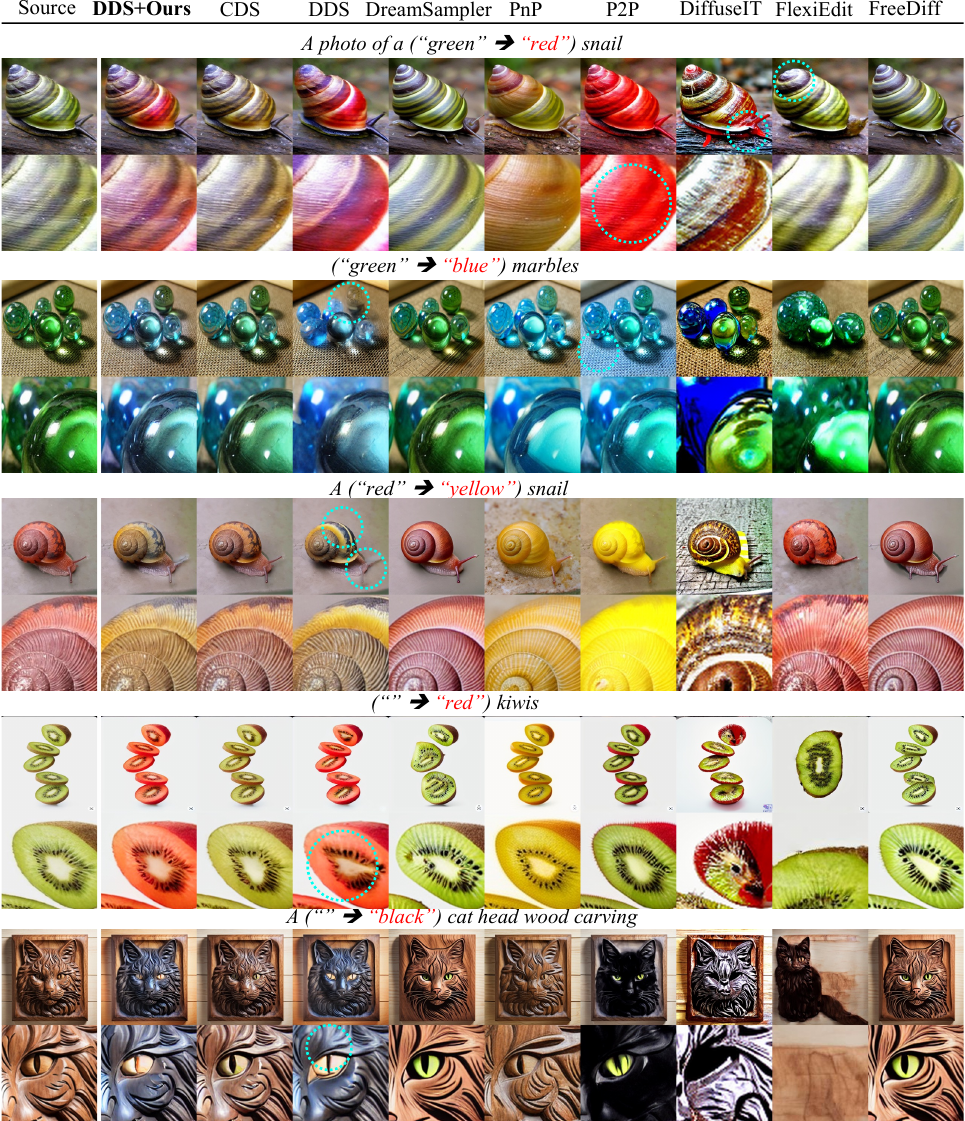} \caption{\textbf{Qualitative results on detail preservation.} Our method effectively preserves detailed textures, such as the texture on a snail's shell and the reflection highlights on marbles. CDS, FlexiEdit, and FreeDiff do not follow instructions well, and other methods distort local details. (Best viewed on a screen when zoomed in)}  \label{supp:fig:2d-low-freq-edit2} \end{figure*}

\noindent \textbf{Additional Results in Detail Preservation.} As shown in Fig. ~\ref{supp:fig:2d-low-freq-edit1} and Fig. ~\ref{supp:fig:2d-low-freq-edit2}, our method preserves high-frequency details, such as intricate wood carvings, thin elements, and textures.

\noindent \textbf{Additional Results in Color Preservation.} As shown in Fig. ~\ref{supp:fig:2d-low-freq-edit2}, baseline methods often fail to maintain source image colors (e.g., orange tones and cat colors), whereas our method achieves better color consistency compared to DDS and CDS.

\begin{figure*}[t] \centering \includegraphics[width=1.0\textwidth, page=6]{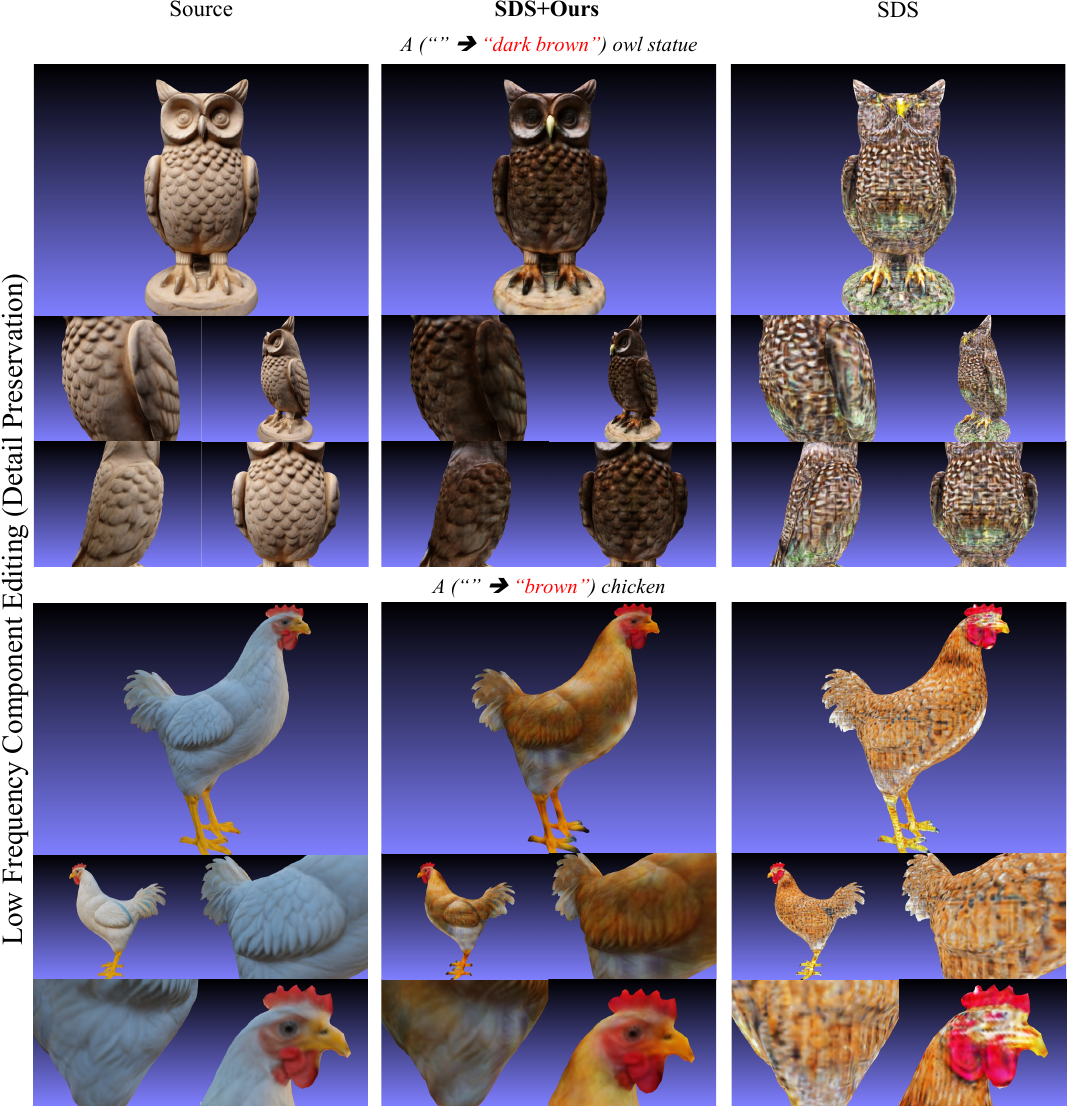} \caption{\textbf{Qualitative results on texture low frequency editing.} Compared to vanilla SDS, which largely ignores the detailed patterns of the original texture, SDS+Ours produces edits that preserve these details, such as the feathers of a chicken. Notably, SDS nearly removes the eyes of the owl and the chicken, as they are high-frequency details, while our method preserves them well. (Best viewed on a screen when zoomed in and also see in attached videos)}  \label{supp:fig:3d-low-freq-edit1} \end{figure*}

\begin{figure*}[t] \centering \includegraphics[width=1.0\textwidth, page=7]{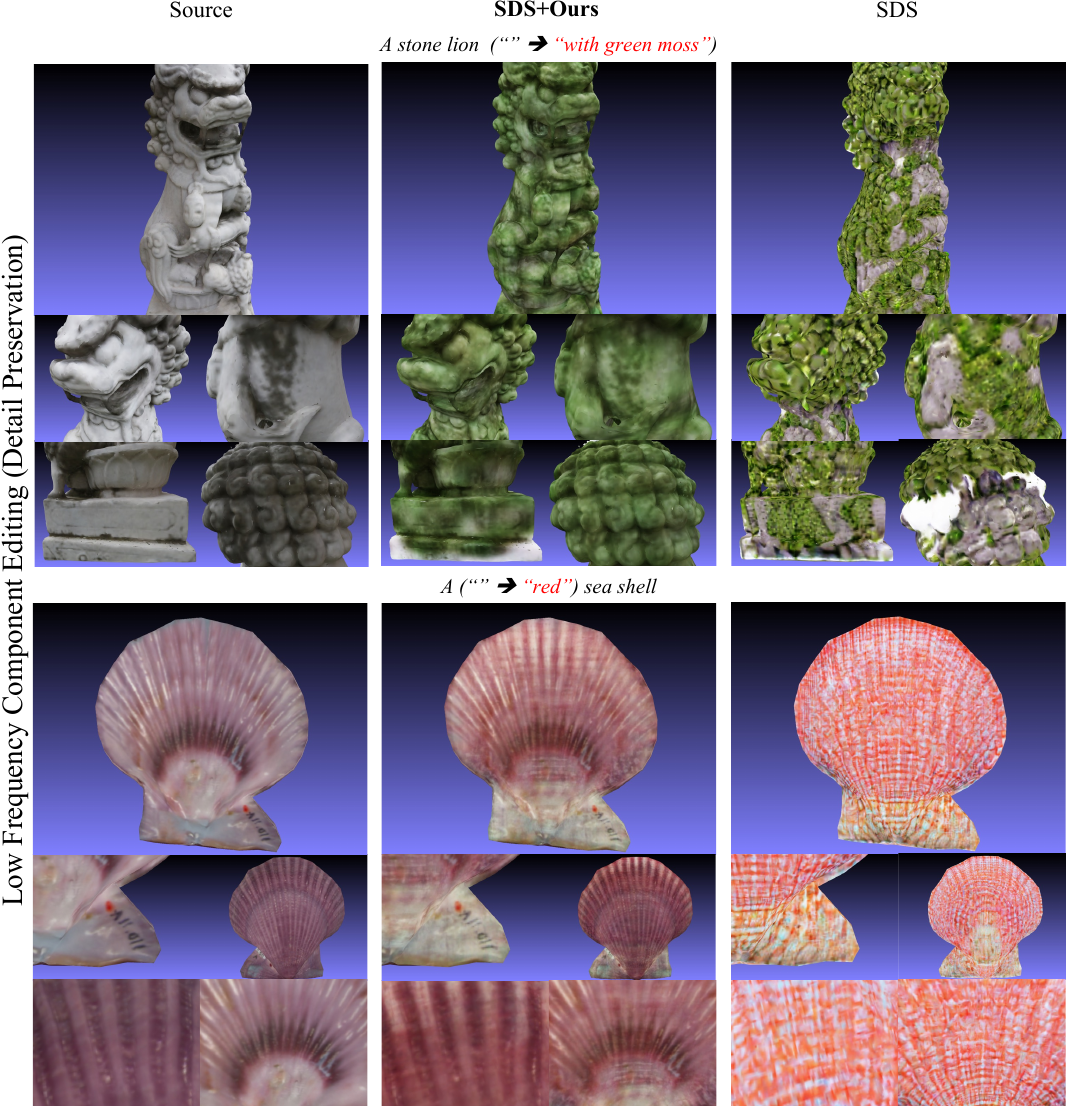} \caption{\textbf{Qualitative results on texture low frequency editing.} Our method preserves high-frequency details, such as the intricate texture on the shell and the fine lines on the stone. In contrast, SDS alone drastically alters these details, resulting in unrealistic-looking objects (as in the stone lion case) or entirely different objects (as in the sea shell case). (Best viewed on a screen when zoomed in and also see in attached videos)}  \label{supp:fig:3d-low-freq-edit2} \end{figure*}

\begin{figure*}[t] \centering \includegraphics[width=1.0\textwidth, page=8]{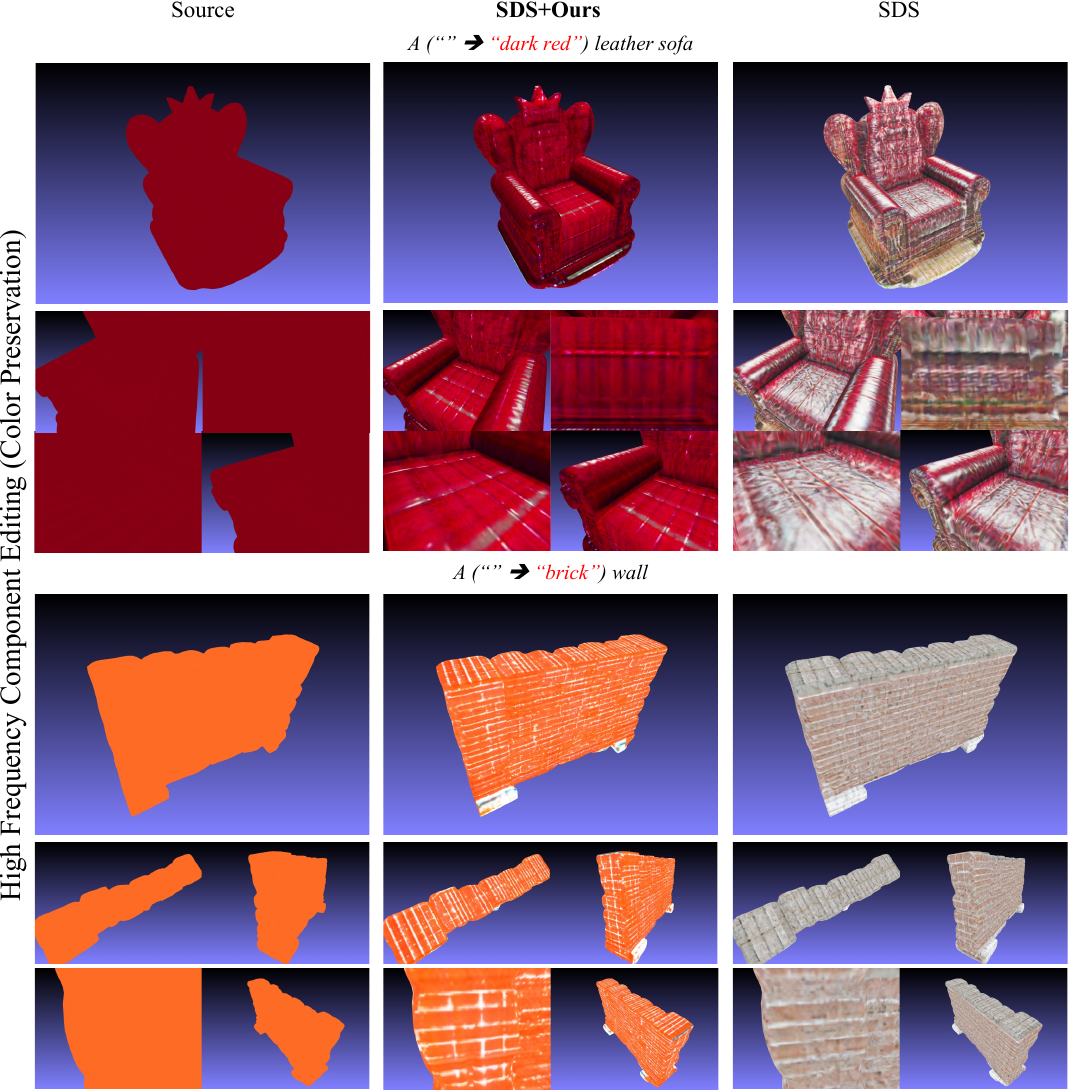} 
\caption{\textbf{Qualitative results on texture high frequency editing.} 
The goal of high-frequency texture editing is to make modifications while preserving the original object's color. As shown, SDS drastically alters the texture, almost entirely disregarding the original texture map. In contrast, our method adjusts high-frequency details while maintaining the original color of the sofa and the wall. (Best viewed on a screen when zoomed in and also see in attached videos)} 
\label{supp:fig:3d-high-freq-edit1} \end{figure*}

\noindent \textbf{Additional Texture Editing Results.} We provide additional texture editing results in Fig. ~\ref{supp:fig:3d-low-freq-edit1}, Fig. ~\ref{supp:fig:3d-low-freq-edit2}, and Fig. ~\ref{supp:fig:3d-high-freq-edit1}. Compared to SDS, SDS+Ours achieves better detail preservation, such as the patterns on an owl's and a chicken's feathers, as well as improved color consistency, such as the color of a sofa. A video is attached to this supplementary material for better viewing quality.

\noindent \textbf{Additional Gradient Visualizations.} We provide additional gradient visualizations for both detail preservation and color fidelity cases in Fig. ~\ref{supp:optimization_matcha} and Fig. ~\ref{supp:optimization_stone}, respectively.

\begin{figure*}[t] \centering \includegraphics[width=1.0\textwidth, page=1]{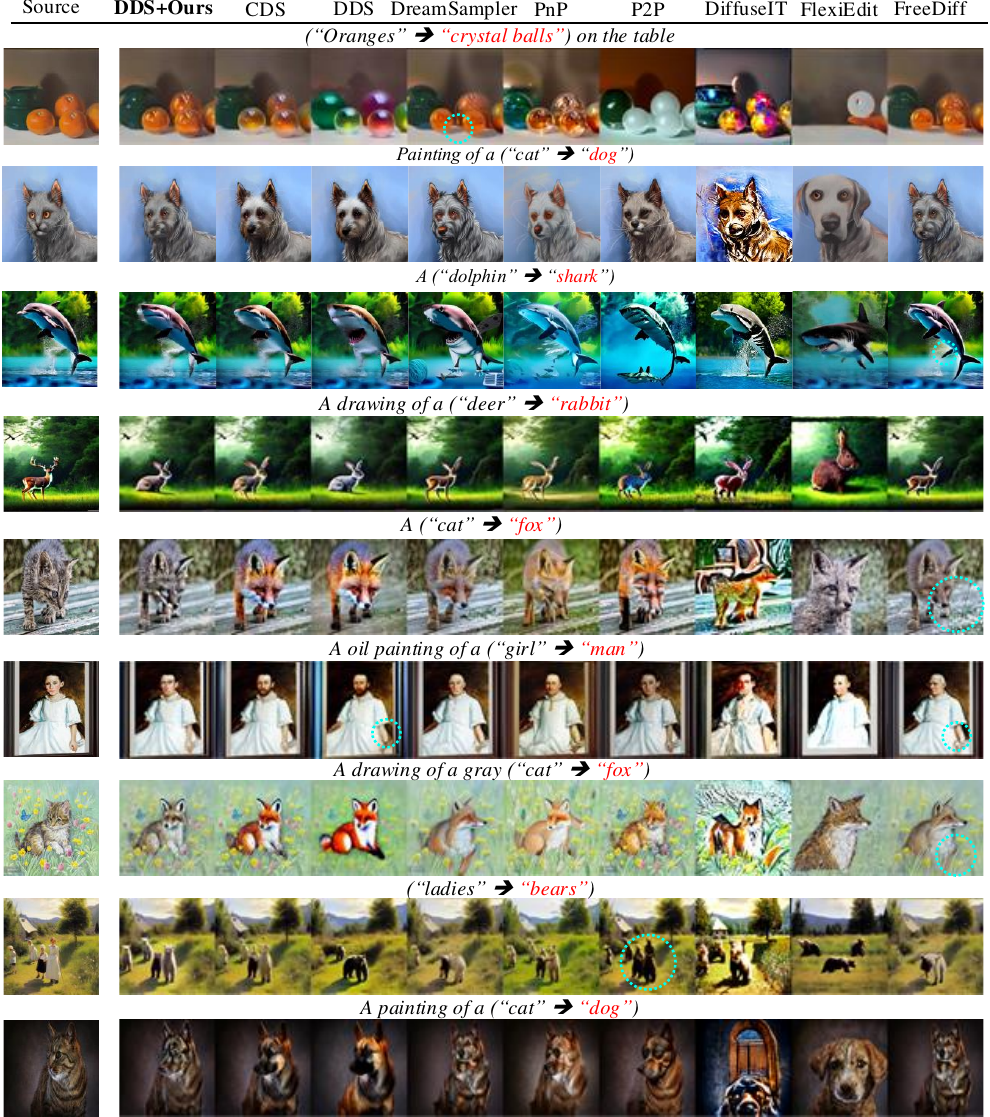} \caption{\textbf{Qualitative results on color preservation.} Our method maintains the colors of the source image, such as the orange color of oranges and the fur color of animals, resulting in edits that not only follow the editing instructions but also appear more consistent with the source image. While FreeDiff also preserves colors, it fails to maintain geometric structures, such as altering the head position when turning a cat into a fox. (Best viewed on a screen when zoomed in)}  \label{supp:fig:2d-high-freq-edit} \end{figure*}

\twocolumn[{
\begin{center}\captionsetup{type=figure}\includegraphics[width=1.0\textwidth, page=9]{figs/figs-supp.pdf} \caption{\textbf{Additional gradient visualization during optimization.} Edit with “A cup of (coffee → matcha)”. The high frequency gradient $\phi_\text{H.F.S.}^1$ and $\phi_\text{H.F.S.}^2$ in DDS distort the detail pattern of latte art. With frequency awareness, our method is able to preserve these details.} 
\label{supp:optimization_matcha}
\end{center}
\begin{center}\captionsetup{type=figure}\includegraphics[width=1.0\textwidth, page=10]{figs/figs-supp.pdf} \caption{\textbf{Additional gradient visualization during optimization}. Edit with ``A (stack of stones $\rightarrow$ Buddha statue)". The low frequency component of $\phi_\text{L.F.S.}$ is changed modified and thus the color of the stone is turned into yellowish. (Best viewed on a screen when zoomed in)} 
\label{supp:optimization_stone}
\end{center}
}]

\end{document}